\documentclass[letterpaper]{article}

\usepackage{arxiv}

\usepackage[utf8]{inputenc} 
\usepackage[T1]{fontenc}    
\usepackage{hyperref}       
\usepackage{url}            
\usepackage{booktabs}       
\usepackage{amsfonts}       
\usepackage{nicefrac}       
\usepackage{microtype}      
\usepackage{graphicx}
\usepackage{doi}

\title{Platform for Situated Intelligence}

\author{ {Dan Bohus, Sean Andrist, Ashley Feniello, Nick Saw, Mihai Jalobeanu}, \\
    \textbf{Patrick Sweeney, Anne Loomis Thompson, Eric Horvitz} \\
    \\
    Microsoft Research\\
	One Microsoft Way\\
	Redmond, WA 98052 \\
	\\
	\texttt{\{dbohus;sandrist;ashleyf;nick.saw;mihaijal;patricsw;annelo;horvitz\}} \\
	\texttt{@microsoft.com} \\
}

\date{}

\renewcommand{\headeright}{MSR-TR-2021-02}
\renewcommand{\undertitle}{}
\renewcommand{\shorttitle}{\textsc{Platform for Situated Intelligence}}

\hypersetup{
pdftitle={Platform for Situated Intelligence},
pdfauthor={David S.~Hippocampus, Elias D.~Striatum},
pdfkeywords={First keyword, Second keyword, More},
}

\begin{document}
\maketitle

\newcommand{\psif}{\emph{\textbackslash psi }}
\newcommand{\psifnospace}{\emph{\textbackslash psi}}


\newcommand{\psistudio}{\textsc{PsiStudio} }
\newcommand{\psistudionospace}{\textsc{PsiStudio}}


\newcommand{\psistore}{\textsc{PsiStore} }
\newcommand{\psistorenospace}{\textsc{PsiStore}}

\begin{abstract}
We introduce Platform for Situated Intelligence, an open-source framework created to support the rapid development and study of multimodal, integrative-AI systems. The framework provides \emph{infrastructure} for sensing, fusing, and making inferences from temporal streams of data across different modalities, a set of \emph{tools} that enable visualization and debugging, and an ecosystem of \emph{components} that encapsulate a variety of perception and processing technologies. These assets jointly provide the means for rapidly constructing and refining multimodal, integrative-AI systems, while retaining the efficiency and performance characteristics required for deployment in open-world settings.
\end{abstract}


\section{Introduction}
\label{sec:Introduction}

Recent advances in machine learning have led to significant improvements on numerous perceptual tasks \cite{Goodfellow-et-al-2016}. For instance, in the span of a decade, error rates on object detection tasks have improved from \textasciitilde 50\% in 2010 to \textasciitilde13\% in 2020 \cite{ImagenetLeaderboard, alom2018history, russakovsky2015imagenet, touvron2019fixing}. Similarly, error rates on conversational speech have dropped dramatically \cite{xiong2018microsoft, saon2017english, SwitchboardLeaderboard}. Large strides have been made in machine translation \cite{wu2016google}, reading comprehension and text generation \cite{brown2020language}, recommender systems \cite{zhang2019deep}, dexterous robot control \cite{andrychowicz2020learning}, and mastering competitive games \cite{schrittwieser2019mastering}.

Despite the steady progress in perceptual and control technologies, current AI models, and larger systems that incorporate them, provide singular, narrow wedges of expertise. A promising pathway to developing more general AI capabilities centers on bringing together and coordinating a constellation of AI competencies. Such integrative approaches can be employed to enable AI systems to perceive key aspects of the physical world and to make inferences across several distinct streams of data, including perceptual signals of the form that people depend on to assess situations and take actions. From mobile robots and cashier-less shopping experiences, to self-driving cars, intelligent meeting rooms, and factory floor assistants, computer systems that operate in the open world need to perceive their surroundings through multiple sensors, make sense of what is going on moment by moment in their environment, and decide how to act in a timely and appropriate manner. 

While many applications of AI will be autonomous, a particularly important, yet challenging opportunity for AI is developing intelligent systems that can collaborate in a natural manner with people. Fluid human-AI interaction will require AI systems to sense, infer, and coordinate with people with the ease, speed, and effectiveness that people expect when working with each other. Another important capability is formulating and leveraging a shared understanding or grounding with people about the task at hand, reminiscent of the shared understandings that people assume when they collaborate with one another. Such human-centered capabilities will hinge on endowing AI systems with multimodal capabilities that enable them to see, listen, and speak, and to understand critical aspects of language, gestures, and the surrounding physical environment.

Setting aside the many inspiring scientific research questions inherent to this vision of the future, constructing \emph{multimodal, integrative-AI systems} that operate effectively in the open world is a very challenging task from an engineering perspective. The engineering challenges arise primarily due to the mismatch between the requirements of the task at hand and the programming languages, infrastructures, and development tools we currently have available. For example, to make sense of the surrounding world, physically situated systems need to integrate information from multiple sensors and leverage skills provided by a heterogeneous set of component AI technologies. Capabilities are required to continually analyze and make inferences under uncertainty from multiple streams of data, considering distinctions about actors and objects and their behaviors over space and time. Yet constructs such as time, space, and uncertainty are not first-order objects in current programming fabrics. The development cycle for today's systems is often data-driven, and involves training or using existing machine learned models, inspecting multiple runs and computing quality metrics, iterative refinement and tuning of parameters, etc. In this context, tools for data visualization, annotation, and analytics can be instrumental during development, yet such tools are often missing, or are spread across multiple ecosystems and not tightly integrated with the rest of the development cycle.

To directly address these engineering challenges, and to provide a solid foundation for development and study of multimodal, integrative-AI systems, we have developed an open-source framework\footnote{Platform for Situated Intelligence is available at \url{https://github.com/microsoft/psi}} called \emph{Platform for Situated Intelligence} \cite{bohus2017icmi, andrist2019hri}. The framework---which we abbreviate as \psifnospace, pronounced like the Greek letter $\psi$---provides a modern infrastructure along with tools that are specifically tailored to the needs of these applications. Concretely, \psif subsumes (1) a \emph{runtime} which provides a parallel, coordinated computation programming model as a basis for writing efficient integrative-AI applications that operate with multimodal streaming data, (2) a set of \emph{tools and APIs} that support and accelerate the development, debugging and maintenance cycle for these applications, and (3) an open \emph{ecosystem of components} which wrap a variety of AI technologies, promoting reuse and further accelerating the pace of development. \psif has been motivated by our team's previous experiences in robotics and situated interaction, including research and development of systems that require fluid and precisely timed coordination with humans\footnote{For a video overview of our situated interaction research, see \url{https://youtu.be/ZHsQYu2X7eY}} (e.g., \cite{bohus2009dialog, bohus2011decisions, bohus2014managing, yu2015incremental, bohus2017study, bohus2019chapter}). However, the framework's reach and use-cases are significantly broader: any application that processes streams of data, and where timing is important, can benefit from the programming models, primitives, and tools provided by \psifnospace.

In the following sections, we provide a broad, comprehensive overview of \psifnospace. We begin by expanding on the points above regarding motivation and goals in the next section, and continue by providing a high-level overview of the framework in Section \ref{sec:Overview}. Next, we focus in turn on each of the three major areas of the framework: we discuss the \psif runtime in Section \ref{sec:Runtime}, followed by the set of available tools in Section \ref{sec:Tools}, and finally the extensible component ecosystem in Section \ref{sec:Components}. In Section \ref{sec:RelatedWork} we discuss related work and situate the \psif framework in relation to other existing infrastructures. Finally, in Section \ref{sec:Conclusion} we summarize the most important aspects of the framework and conclude with a call to action.

\section{Motivation and Goals}
\label{sec:Motivation}

Developing and maintaining multimodal, integrative-AI systems brings to the fore multiple engineering challenges. First, the \emph{multimodal} nature of these systems raises numerous issues. Systems with multimodal capabilities process data of different types, generated by a wide array of sensors, such as audio, visual, depth, inertial measurement units, and lidars. The data is continuously streaming, often at high-bandwidth, and must be processed under latency constraints. For efficiency reasons, computations must often be parallelized or executed asynchronously, yet they must also be closely coordinated, as information from multiple modalities must be fused to support accurate inferences and to make optimal decisions.

Additional challenges stem from the \emph{integrative} nature of these systems: they often need to weave together and carefully orchestrate a heterogeneous array of AI components and technologies. The components use different techniques to process different types of data, which may arrive at different frequencies and latencies. Some components may execute locally, on device, whereas others may run as services in the cloud. Some components may be human authored, producing unit-testable deterministic results, whereas others may be machine-learned from data, producing stochastic outputs. Regardless, their operation must be coordinated in real-time and finely tuned in order to create a well-functioning end-to-end system.

Unfortunately, the typical programming languages, environments and tools used in developing these systems are not well tailored to support the physically situated, multimodal and integrative-AI aspects of the task at hand, introducing still more difficulties for developers. For example, constructs such as time, space, and uncertainty play a fundamental role in physically situated applications, yet these are not first-class objects in any of the generic programming languages we use today. In the absence of libraries that provide for these abstractions, development often starts by first building the requisite infrastructure for representing and reasoning about such constructs. The abstractions built often end up serving the specifics needs of the task at hand, which can hinder generalization and reuse across applications and domains.

Important shortcomings can also be identified with respect to the debugging, development, and maintenance tools available today. Given the asynchronous execution model often employed by these applications, coupled with their reliance on temporal streaming data, traditional debugging techniques such as tracing and breakpoints are not efficient. Instead, it is essential to be able to visualize and closely inspect the data flowing through the application over time. The development process is often data driven, involving iterative cycles of parameter tuning and optimization, and these operations are also not well supported by existing infrastructures and tools. Numerous challenges also arise with testing in complex, integrative, end-to-end systems. Some of the challenges are due to the lack of reproducibility arising from asynchronous, multi-threaded execution; other challenges arise from the use of data-driven, machine-learned models that produce stochastic, non-deterministic results, which propagate in non-linear ways through the rest of the system \cite{nushi2017on}.

Overall, a large impedance mismatch exists between the generic software development infrastructures and tools available, and the novel requirements brought forward by multimodal, integrative-AI systems. As a result, developing such systems incurs a high engineering cost. Since research in this space often involves constructing end-to-end prototypes that act as test-beds for experimentation, it is not surprising that progress in this area has been slow, despite significant advances in individual AI component technologies. \emph{Platform for Situated Intelligence} (\psifnospace) was developed to alleviate these issues, lower the engineering costs, and promote and accelerate research in this space.

\section{A High-Level Overview}
\label{sec:Overview}

Platform for Situated Intelligence is built on \textsc{.NET} standard. The framework is open-source, cross-platform, and retains the ease of programming and modern affordances of a managed programming environment, such as type safety and memory management, while aiming to address the stringent performance needs of multimodal, integrative-AI applications. The framework consists of:

\begin{itemize}
\item \textbf{\textit{a runtime}} that provides the core infrastructure for working with temporally streaming data;
\item \textbf{\textit{a set of tools}} that enable debugging, data visualization, annotation and analysis;
\item \textbf{\textit{an ecosystem of components}} that wrap a variety of technologies.

\end{itemize}

\noindent At the high level, a \psif application consists of a graph, or \emph{pipeline} of \emph{components} that communicate with each other via temporal \emph{streams} of data. \psif applications are authored by connecting together multiple \psif components, and leveraging the existing tools to support and accelerate development. Below, we provide a short, high-level overview of the runtime, tools, and components, and explain with a simple example how they together support the development of multimodal, integrative-AI applications. The rest of the paper discusses each of these areas in more detail.

\subsection{Runtime, Tools, and Components}

\textbf{Runtime.} The \psif\textit{runtime} provides a simple, modern, type-safe programming model for authoring and executing programs that leverage parallel, coordinated computation in multi-core environments. More specifically, the runtime implements a publish-subscribe architecture \cite{kortenkamp2016robotic}, where computation happens concurrently in a graph of components (nodes) that are connected to each other via streams of data (edges). Message delivery is orchestrated by a runtime scheduler, which aims to efficiently utilize available computational resources and ensure progress. Time is a first-order construct in the framework: the runtime provides abstractions for reasoning about time and for synchronizing data streams, enables latency-awareness in applications from the bottom up, and provides levers that enable application developers to control how the computation degrades under load, i.e., when components cannot keep up. In addition, the runtime provides several other important affordances designed to directly target and simplify the development of multimodal, integrative-AI systems: automatic serialization and high-throughput logging of data streams into persistent data stores, isolated execution via automatic data cloning, data replay, remoting, etc. We discuss the various aspects of the runtime in more detail in Section \ref{sec:Runtime}.

\textbf{Tools.} The framework also provides a \emph{set of tools} that support and accelerate data-driven debugging, testing and tuning scenarios. Specifically, \psif includes \emph{Platform for Situated Intelligence Studio} (abbreviated \psistudionospace) --- a multimodal, temporal-data visualization, annotation and analytics tool illustrated in Figure \ref{fig:PsiStudio-Main}. The tool provides an open, extensible set of visualizers which can be composited in complex layouts to inspect data. The tool supports easy navigation and inspection of data across time, and across collected data stores. In addition to enabling offline visualization for streams persisted by \psif applications, \psistudio also allows for live data visualization, while the application is running. \psif also provides data processing APIs that enable batch processing, re-running pipelines over existing data, data analytics and data extraction for machine learning. While many of these features are integrated for easy access via the \psistudio user interface, they are also all available programmatically, via APIs, as well as from a command-line tool. We review \psistudio and the set of APIs in more detail in Section \ref{sec:Tools}.

\textbf{Components.} Finally, the \psif framework also includes a \emph{set of components} which can be easily connected to each other and provide the basis for rapidly prototyping applications. The set of components currently available in the \psif repository focuses on multimodal, audio-visual sensing and processing technologies. It includes sensor components for cameras, microphones, depth sensors, etc., a variety of audio, image, and speech processing components, as well as wrapper components that enable running machine-learned models in ONNX format, or that enable using Azure Cognitive Services, and so forth. We plan to continue to extend this set of components, with a focus on embodied, physically-situated interaction. New components can be easily developed and added. We expect that the existing set of components will grow into a larger ecosystem through community contributions and in the process further lower the barrier to entry for developing integrative-AI applications.

\subsection{A Simple Example}

To briefly illustrate the fundamentals of the \psif framework, and the role that the runtime, tools, and components play in developing \psif applications, we begin with a simple example. 

Consider the end goal of developing a robot that interacts with people via natural language in an open, noisy space such as a building lobby. One of the basic problems such a robot would face is that of \emph{voice-activity-detection} (VAD), i.e., determining when someone is speaking. Traditional approaches for VAD rely on audio processing, and attempt to determine whether at any point in time the acoustic signal captured by the microphone contains speech or silence. In an open environment however, background noises and speech produced by other distant bystanders can lead to a significant number of false-positives, i.e., situations when the robot thinks someone is talking to it but in reality no one is. Suppose we want to reduce such false-positives and increase the overall robustness of VAD by leveraging visual information. Specifically, we will aim to detect lip movements from people in front of the robot and consider that speech is being produced only during periods when voice activity is detected in the audio \emph{and} a participant's mouth is open. While real-world applications built on \psif are often significantly more complex, this \emph{multimodal VAD} example is sufficient to illustrate some of the main concepts in the \psif framework.

\begin{figure}[!b]
    \centering
    \includegraphics[width=0.8\textwidth]{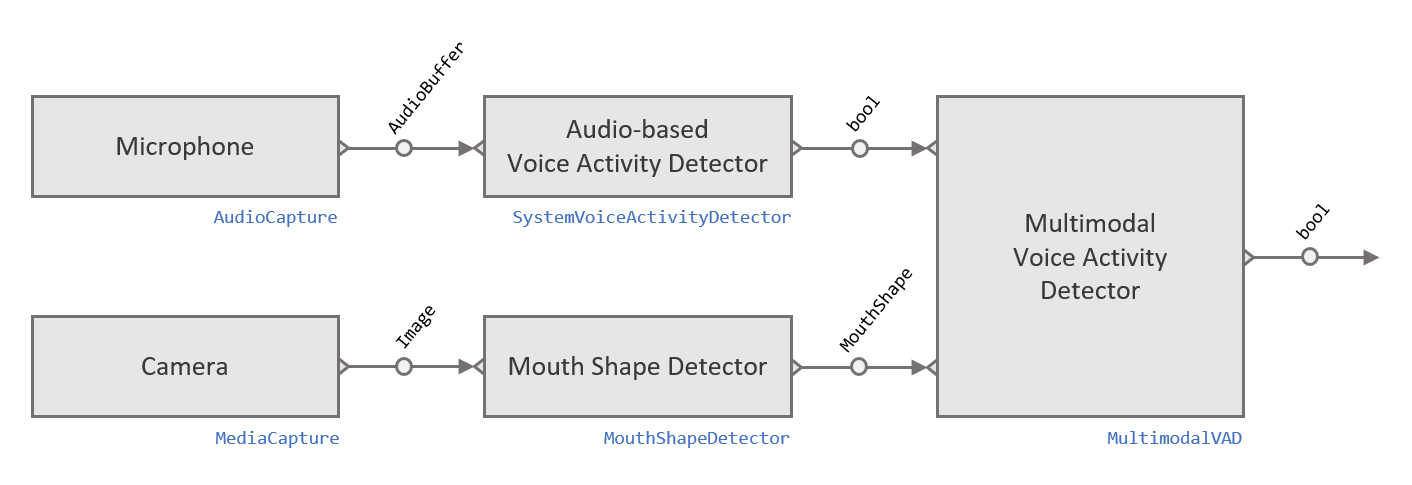}
    \caption{A simple, hypothetical application pipeline which fuses information from camera and microphone source components to accomplish multimodal voice-activity-detection.}
    \label{fig:SimpleAppPipeline}
\end{figure}

A \psif application pipeline typically contains a few \emph{source} components that generate streams of data, such as cameras, microphones, depth sensors, eye trackers, etc., and various downstream processing components. For instance, the hypothetical pipeline from the multimodal VAD example described above is shown in Figure \ref{fig:SimpleAppPipeline}. It contains two source components: a microphone and a camera, and three processing components: an audio-based voice-activity-detector, a mouth shape detector, and a fusion component that brings together the information from the acoustic and visual channels and creates the desired, more robust voice-activity-detection signal.

\begin{figure}
    \centering
    \fbox{\includegraphics[width=0.7\textwidth]{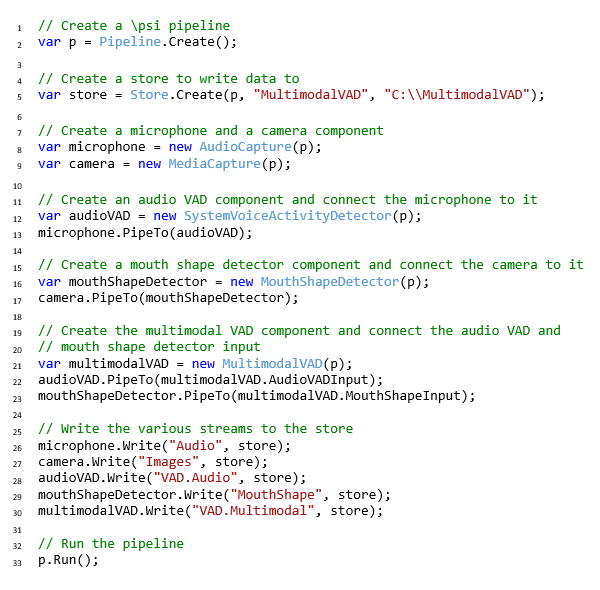}}
    \caption{Example code for a simple \psif application.}
    \label{fig:SimpleAppCode}
\end{figure}

The \psif repository currently includes a wide array of sensor and processing components, and the runtime provides primitives and APIs that enable authoring new components. We discuss the techniques for constructing new components in Section \ref{subsec:Runtime_Components}, and review the set of currently available components later on, in Section \ref{sec:Components}. Wiring together components into pipelines is straight-forward, as the code shown in Figure \ref{fig:SimpleAppCode} demonstrates: components are created by instantiating new corresponding objects (see lines 8, 9, 12, 16, 21 in Figure \ref{fig:SimpleAppCode}), and connections between components are created via the \texttt{PipeTo} stream operator (see lines 13, 17, 22, 23).

The streams in a \psif application are strongly typed. For instance, the microphone (\texttt{AudioCapture}) component produces a stream of \texttt{AudioBuffer} objects, and the audio-based voice activity detector (\texttt{SystemVoiceActivityDetector})  expects a stream of such objects as input, and produces a stream of \texttt{bool} instances. The data flowing through the streams can be easily persisted to a data store on disk, via a simple \texttt{Write} operator (see lines 26--30 in Figure \ref{fig:SimpleAppCode}).

\psistudio enables multimodal data visualization, debugging, annotation and analysis over persisted data stores. Developers can construct complex visualization layouts to view temporal data, navigate over time, inspect data values, and leverage additional analytics and annotation capabilities. \psistudio provides a wide, extensible array of visualizers, and enables both offline visualization, i.e., inspecting data from a persisted store, as well as live visualization, i.e., visualizing the data streams while the application is running. We review the \psistudio tool in more detail in Section \ref{sec:Tools}.

Once the pipeline is constructed, it can be executed by invoking the \texttt{Run()} method (see line 33 in Figure \ref{fig:SimpleAppCode}). The \psif runtime is in charge of running the pipeline, and controls message delivery across the various streams that connect the components. The runtime leverages multiple threads and coordinates the asynchronous execution of the various components in the pipeline. In the process, it aims to create efficiency gains via pipeline parallelism in multi-core architectures. Component developers are however insulated from the challenges that often arise with concurrent, asynchronous programs (refer to section \ref{subsec:Runtime_Components} for additional details). They can also control and fine tune the behavior of the pipeline under load, by essentially specifying where and when messages can be dropped, or when sources can be throttled---i.e., signaled to slow down in producing messages---in order to keep up the pace. Finally, as will be covered shortly, the primitives and operators provided by the runtime simplify data synchronization and fusion, and enable coordination, as well as data replay scenarios. We review these aspects in more detail in the next section.

Altogether, the runtime, tools and components available in the framework simplify and streamline development efforts both at the level of the \emph{component developer}, i.e., the person in charge of authoring new components, and at the level of the \emph{application developer}, i.e., the person in charge of constructing applications that use such components. The same developer can alternate between both roles, using off-the-shelf components provided by the core framework or available in the community in order to assemble their application, while also occasionally writing new components to target application-specific functions. In the next three sections, we describe in turn the \psif runtime, tools, and components, as they relate to these two different roles. 

\section{Runtime}
\label{sec:Runtime}

The \psif runtime infrastructure provides the programming and execution model for parallel, coordinated computation over components connected by time-aware data streams. The runtime is implemented on \textsc{.NET} Standard and is cross-platform compatible. Overall, it was designed to retain the modern affordances and software engineering benefits of a managed programming language like C\#, such as type safety and memory management, while still targeting the demanding performance requirements of multimodal, integrative-AI applications.

In the following sub-sections, we review the \psif runtime in more detail. We begin by discussing \emph{streams} and \emph{components}, which are core primitives in the runtime and represent the edges and nodes of the computational graph. Then, in Section \ref{subsec:Runtime_PipelineExecution} we discuss the core runtime mechanisms that control the execution of the pipeline. We then turn our attention back to components: we discuss the stream operators provided as part of the \psif runtime that enable transforming, synchronizing, and fusing streams of data in Section \ref{subsec:Runtime_StreamOperators}, and the means for hierarchically organizing components in Section \ref{subsec:Runtime_CompositeComponents}. Finally, in Section \ref{subsec:Runtime_RemotingInterop} we briefly review remoting and interop capabilities.

\subsection{Streams}
\label{subsec:Runtime_Streams}

Streams are a core primitive in the \psif runtime: they correspond to edges in the application pipeline and are used to carry data between components. 

\psif streams are implemented via \textsc{.NET} generics: a stream carrying messages of type \texttt{T} is represented by an \texttt{IProducer<T>}. A stream of type \texttt{T} can only be connected to a component that has an input of the same type, and the typed nature of \psif streams enables static type checking of the application pipeline. 

\begin{figure}
    \centering
    \includegraphics[width=\textwidth]{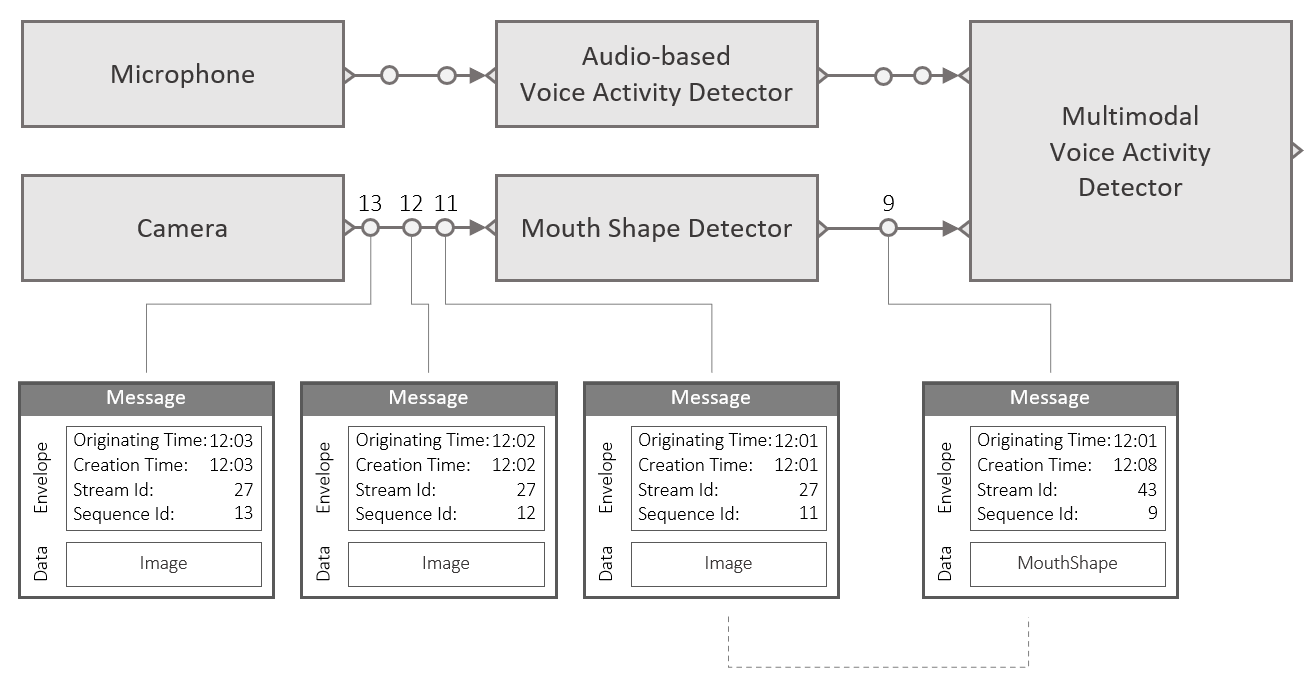}
    \caption{Messages on a \psif stream. An image message (sequence ID: 11) has been emitted by the \texttt{Camera} source component (stream ID: 27) at 12:01. That message was then processed by the \texttt{Mouth Shape Detector} component and emitted (on stream ID: 43) as a new message (sequence ID: 9 on the output stream from the mouth shape detector) at 12:08, but retains the same originating time of 12:01. This message thus has a known latency of seven seconds. (The timestamps in this illustrative example are not realistic and rounded to the second for clarity purposes.)}
    \label{fig:SimpleAppMessages}
\end{figure}

Time is a first-order construct in \psif streams. When components emit data on streams, each piece of information that is posted is wrapped with an \emph{envelope} which contains important timing and sequencing information. Together, the data contents and the envelope form the \emph{message} that is passed along the stream from one component to another.  Messages flowing in a stream of type \texttt{T} are of type \texttt{Message<T>}, containing a \texttt{Data} member, which captures the data contents, of type \texttt{T}, and an \texttt{Envelope} member which contains auxiliary information, as illustrated in Figure \ref{fig:SimpleAppMessages}. 

The envelope for each message contains information about the stream id and order of the message in the sequence, as well as two different timestamps: the message \emph{creation-time} and the message \emph{originating-time}. The message \emph{creation-time} captures the time at which the message was created by the component that generated it. In contrast, the \emph{originating-time} is assigned once, by the source component (e.g., by the camera, microphone, etc.), and is carried forward through the pipeline. The \emph{originating-time} for a message therefore represents \emph{the moment in time to which the message corresponds in the real world}. All consecutive messages traveling on a particular stream must have strictly increasing creation and originating times.

For instance, in the example shown in Figure \ref{fig:SimpleAppMessages}, three successive images, with sequence IDs 11, 12, and 13, are captured by the camera at times 12:01, 12:02, and 12:03 respectively. (The rounded timings presented in this example were artificially chosen for clarity of exposition, but in reality the runtime allows for obtaining high resolution time information.) In this case, these three messages have matching originating and creation-times---we assume here that no significant time elapses between the moment an image is captured and when the message is created. As the images are passed downstream, the mouth shape detector component will process them one by one, and post the corresponding mouth-shape results on its output stream. Since the visual analysis for detecting mouth shapes takes some time, these messages will be created later, and therefore have a later creation-time. For instance, in the illustration from Figure \ref{fig:SimpleAppMessages}, the mouth shape detector generates a result for the first image only at creation-time 12:08. Nevertheless, the envelope for this output message also contains the originating-time from the image that generated it, i.e., 12:01. This propagation of originating-time information occurs throughout the entire application pipeline. As a result, each message inherently carries with it latency information: for instance, the first mouth shape result we have discussed has a latency of 7 seconds, as reflected by the delta between its creation-time and originating-time.

The \emph{latency-aware} nature of \psif streams provides a number of important benefits and enables several key scenarios. Each component in the application pipeline can inspect the delay of each incoming message with respect to the world, and can make decisions based on that information. For example, a component may choose to drop---i.e., not to process---an input message if that message is already too delayed with respect to the world. In another scenario, a component may use latency information to extrapolate and construct an estimate of the current value given the time elapsed since the message originated in the world.

The availability of originating-times and latency information also enables stream fusion based on when events actually happen in the world, as opposed to based on when the messages arrive at the component. For instance, the audio-based results and mouth shape results in the example from Figure \ref{fig:SimpleAppMessages} arrive at the multimodal voice activity detection component at different latencies, because visual processing for detecting mouth shapes takes a different amount of time than audio processing for detecting speech. However, because it has access to originating-times, the component can fuse the messages on the two streams correctly, based on when the events happened in the world. For instance, the mouth shape message with originating-time 12:01 can be fused with a voice activity message with the same 12:01 originating-time, regardless of how long it took to compute each of these messages. The originating-times act as a basis for \emph{reproducible fusion}, i.e., fusion where the \emph{results are independent of the compute delays} in the pipeline. The \psif framework provides a rich set of stream operators that simplify this kind of temporal fusion, merging, sampling, and synchronization operations; we review them in more detail later on, in Section \ref{subsec:Runtime_StreamOperators}.

The \psif runtime also has access to message timing information and is able to make scheduling decisions based on that information. For example, the runtime favors delivery and processing of older messages to generally keep the overall pipeline running at low latency. Furthermore, the \psif runtime allows application developers to configure \emph {delivery policies} on streams such that messages are dropped if their latency is above a specified threshold. Finally, the timing information is also persisted when logging streams to disk. This in turn enables replay scenarios, where streams are constructed and published by reading from logs, and the real-time execution of the pipeline can be accurately simulated. We elaborate on these aspects in more detail in Section \ref{subsec:Runtime_PipelineExecution}, where we discuss scheduling and other core runtime mechanisms.

\subsection{Components}
\label{subsec:Runtime_Components}

While streams correspond to the edges in the application pipeline, \emph{components} are the nodes where the computation actually happens. In general, components receive zero or more input streams, and can produce zero or more output streams. 

\begin{figure}[!b]
    \centering
    \fbox{\includegraphics[width=0.8\textwidth]{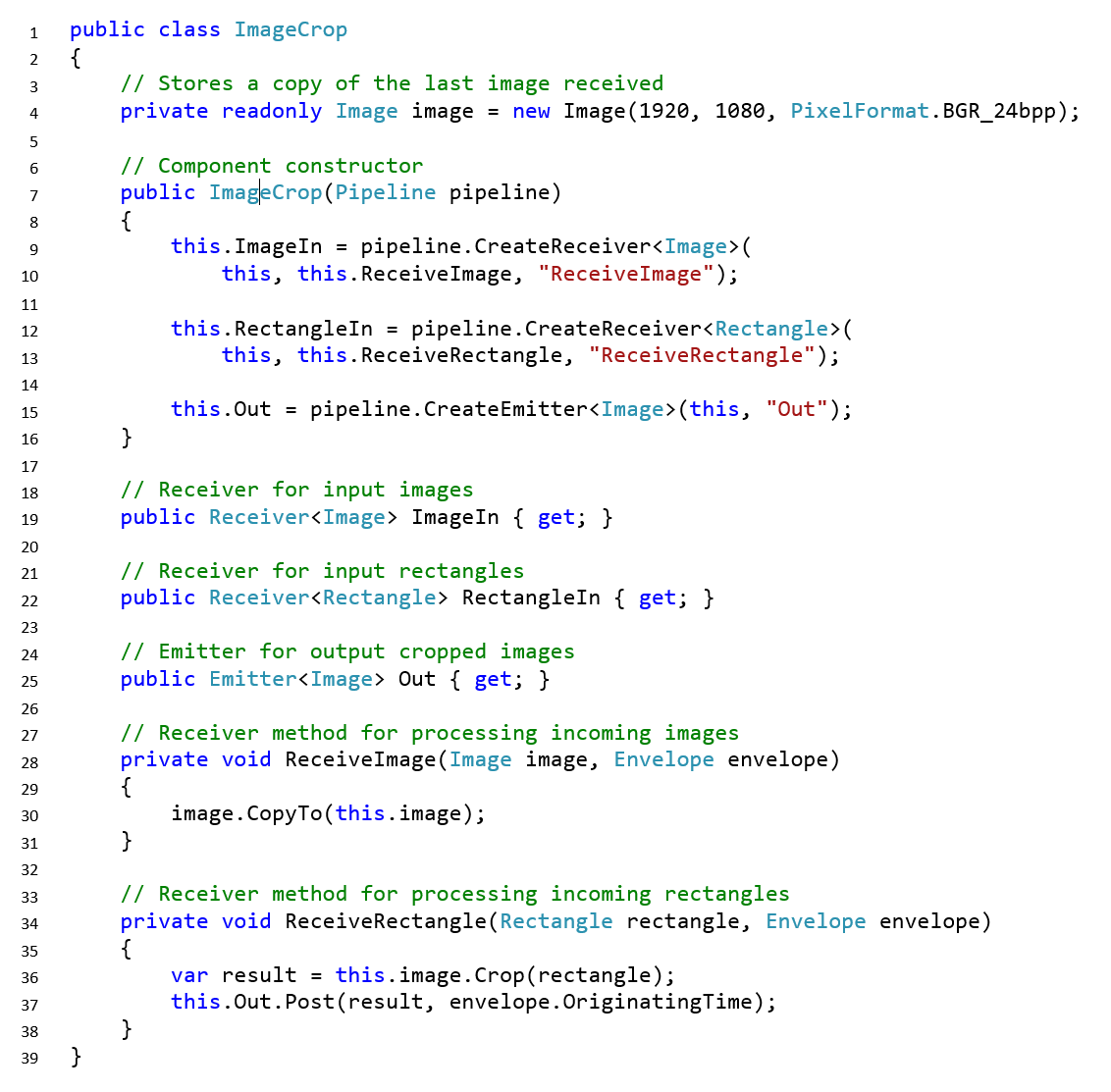}}
    \caption{Example code for a simple \psif component that performs a crop operation with input streams of images and rectangles.}
    \label{fig:CodeComponent}
\end{figure}

In this section, we describe the fundamental anatomy of a basic \psif component and briefly illustrate how developers can write their own components. Figure \ref{fig:CodeComponent} shows a simple example component called \texttt{ImageCrop}. As the name implies, this component performs a crop operation on a stream of incoming images. The component has two input streams, one containing images and one containing rectangles. For every rectangle received, it crops the latest image it received and publishes the resulting cropped image on the output stream. This is a minimal and somewhat artificial example that could be optimized and improved in a number of different ways, but is sufficient to illustrate the basics of writing \psif components. 

The pattern for authoring \psif components is straightforward: the author writes a new C\# class which contains a \emph{receiver} for each of the component's inputs and an \emph{emitter} for each of its outputs. In the example in Figure \ref{fig:CodeComponent}, see lines 19 and 22 for the image and rectangle receivers, and line 25 for the output emitter. The component's constructor takes in as a parameter the \texttt{Pipeline} object that this component will become part of, and constructs the receivers and emitters using an API provided by this pipeline object (see lines 9 through 15). When constructing a receiver, the author specifies a \emph{receiver method} which will be used to process incoming messages from that input stream---in this case \texttt{ReceiveImage} for images, and \texttt{ReceiveRectangle} for rectangles. As we have alluded to in the previous section, the receiver method has access not only to the data, but also to the entire message envelope containing the timing information (although in this example, that timing information is not used.) Upon receiving the image, this component simply copies it into local storage, and upon receiving a rectangle it crops the stored image and posts the result on the output stream (lines 36--37).

We have seen that \psif applications are written by connecting multiple such components via streams of data to form a computation graph, or pipeline. The example \texttt{ImageCrop} component discussed above is a \emph{processing component}, i.e., a component that receives one or more (in this case two) input streams, and produces zero or more (in this case one) output streams. \emph{Source components} are another type of components that only produce outputs without having any input streams. Source components generally model sensors, such as cameras or microphones, and act as the sources of data from the outside world into the pipeline. A source component can have one or more emitters that it posts messages on. A number of additional considerations regarding the authoring of source components are presented in the framework documentation \cite{PsiWebsite_WritingComponents}.

Finally, as we shall discuss shortly, the \psif runtime provides APIs that enable constructing \emph{composite components} hierarchically, by leveraging and connecting existing lower-level components. However, before we can explain composite components, it is important to understand a bit more about the \psif runtime and how exactly a pipeline of components is actually executed.

\subsection{Core Runtime Mechanisms}
\label{subsec:Runtime_PipelineExecution}

As we have already discussed, \psif applications consist of computation graphs composed of multiple components connected via streams of data, using a publish-subscribe \cite{kortenkamp2016robotic} message-passing architecture. Component receivers can subscribe to the emitters of other components, with each receiver subscribing to at most one emitter; multiple receivers can however be subscribed to a single emitter. Once the pipeline is running, messages posted to an emitter will be passed along to all of its subscribed receivers. Computation is performed inside the encapsulated components, in response to messages arriving at their receivers. Each component can maintain individual state information, and as we shall shortly see, no lock-based synchronization is required. 

The pipeline and its component connections are specified in the procedural code written by the application developer, as shown in Figure \ref{fig:SimpleAppCode}. Once the \texttt{Run()} method is invoked, the \psif runtime takes charge of actually executing the pipeline. In the process, the runtime aims to maximize resource utilization and leverage pipeline parallelism. The runtime is also designed to insulate component developers from many challenges that typically arise in concurrent execution environments. To better understand how \psif applications run, and the various affordances the runtime provides to application and component developers, we dive deeper in the next subsections into the core principles and inner mechanisms of the \psif runtime. 

\subsubsection{Scheduling, Delivery Queues, and Delivery Policies}
\label{subsubsec:Runtime_Scheduling}

During execution, the \psif runtime uses a \emph{scheduler} that leverages multiple threads to deliver messages and execute the components in the pipeline concurrently. As new messages are generated, the runtime is in charge of scheduling each component's receiver methods for execution. 

\begin{figure}[!t]
    \centering
    \includegraphics[width=0.8\textwidth]{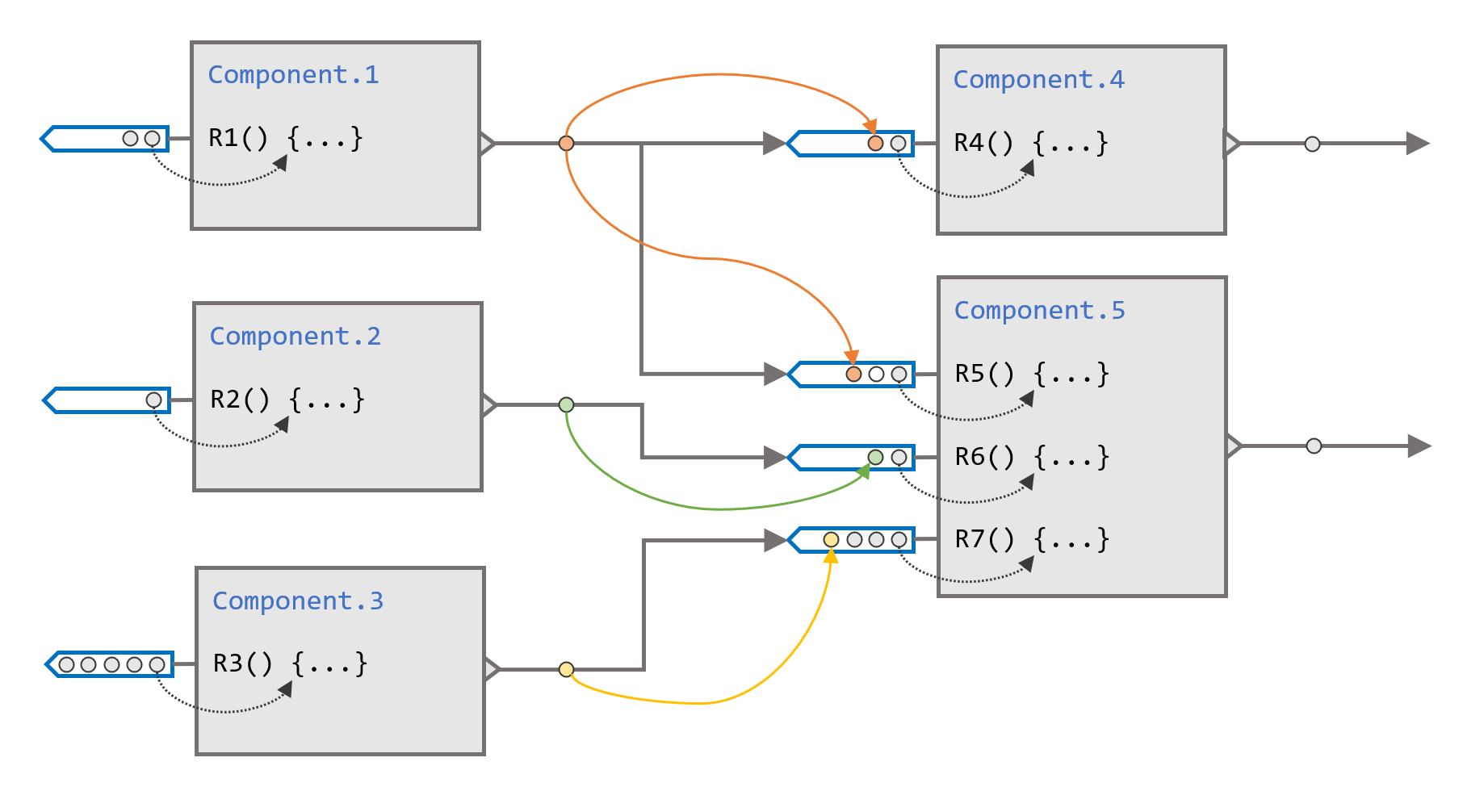}
    \caption{Pipeline with illustrated delivery queues for each receiver (blue outline); colored arrows show enqueueing of messages into the delivery queues---the newer messages in the delivery queues are on the left side; dotted black arrows show message dequeueing and invocation of receiver method;}
    \label{fig:DeliveryQueues}
\end{figure}

In general, messages posted on a component's emitter do not immediately trigger the downstream, subscribed receiver methods. Instead, each receiver method has an associated \emph{delivery queue}. When a message is posted on an emitter connected to that receiver, a copy of the message is added to each of the delivery queues for its subscribed receivers (see Figure \ref{fig:DeliveryQueues}.) For each message, the runtime ensures that a corresponding thread pool work item will be created to pick up that message and invoke the receiver method. This work item will generally execute at a later time, and most likely on a different thread. In general, the runtime prioritizes these work items so that earlier messages (in originating-time) will typically be processed before later ones, allowing the pipeline to leverage the available computational resources to continually make forward progress.

Delivery queues act therefore as a buffer at the entrance of each receiver. It is possible that at the time a new message is posted on an emitter, the delivery queue for its subscribed receiver already contains one or more messages waiting to be delivered. In the default case, the new message, which necessarily has a later originating time than other messages already in the queue, is added to the queue, and all enqueued messages will eventually be processed. This particular behavior is however not always desirable. Consider for instance the case where a receiver method is slow, and does not keep up with the pace at which messages arrive from the emitter it is subscribed to. In this case, the messages will be enqueued at a faster pace than they are dequeued, and as a result, the delivery queue will continue to grow in size, and the corresponding downstream messages created by that receiver method will have increasingly larger latencies. In essence, the component will lag farther and farther behind over time. This behavior can arise not only if the receiver method is slow, but also more generally if there are not enough computational resources available to keep up with incoming messages---the receiver method itself may not be slow, but the CPU load on the machine might be large due to other components, or other processes running on it.

The \psif runtime provides a mechanism that enables application developers to control how the application behaves when the amount of available compute resources changes. Specifically, when creating a connection between an emitter and a receiver via the \texttt{PipeTo} operator, the application developer can specify a \emph{delivery policy}, which defines the behavior of the corresponding delivery queue. The default delivery policy, called \emph{Unlimited}, implements the default behavior described above: all messages are enqueued and will eventually be processed; if new messages arrive faster than they are being processed, the delivery queue can grow to an effectively unlimited size.

At the opposite end of the spectrum, a delivery policy called \emph{LatestMessage} limits the maximum size of the delivery queue to one item: the most recently received message. If a new message arrives while the queue already has an existing message waiting to be delivered, this existing message is dropped and replaced with the new incoming message. With this delivery policy, if the receiver method takes a long time to execute, messages will be dropped from the delivery queue, and when the receiver method becomes available again, only the most recent message will be available to be processed.

While the \emph{Unlimited} and \emph{LatestMessage} delivery policies described above represent two opposite ends of a spectrum, the runtime allows application authors more fine-grained control in customizing delivery policies. For example, application authors can create \emph{queue-size-constrained delivery policies}, where the delivery queue is allowed to grow to a maximum specified size, and \emph{latency-constrained delivery policies}, where the queue is allowed to grow to a maximum specified latency, and older messages that exceed that latency are dropped. The runtime also allows developers to specify policies that take into account the actual contents of the message and guarantee delivery for certain messages. For example, a policy of the form: ``drop image messages with latency larger than 500ms, except if the image is smaller than 200x200 pixels'' can be easily constructed.

In addition, the runtime also supports \emph{throttling} policies, where an upstream component is throttled to slow down its production of new messages when a downstream delivery queue becomes full. Finally, \emph{synchronous} delivery policies are also available---in this case an attempt is made to immediately execute the receiver method on the same thread, when the message is posted on the subscribed emitter, rather than enqueueing the message and generating a corresponding work item. A full description of the set of delivery policies is available for the interested reader at \cite{PsiWebsite_DeliveryPolicies}.

\subsubsection{Receiver Exclusivity}

We have seen that the runtime uses multiple threads to schedule and execute component receiver methods. Apart from generally prioritizing messages with earlier originating times, the \psif runtime obeys another property in this process, which we refer to as \emph{receiver exclusivity}: the receiver methods implemented by a given component are always executed exclusively with respect to each other. While the receivers of different components may be executed concurrently, the runtime automatically enforces exclusivity between the receivers of the same component. For instance, in the example illustrated in Figure \ref{fig:DeliveryQueues}, the receivers \texttt{R1}, \texttt{R2}, \texttt{R3}, \texttt{R4}, and \texttt{R5} may execute concurrently on different threads. However, the runtime guarantees exclusivity between \texttt{R5}, \texttt{R6}, and \texttt{R7} as these receivers belong to the same component. 

This within-component receiver exclusivity property is important, as it enables component-level \emph{state protection}. Developers can design and author components as if they were executed in a single-threaded environment, without concerns about state safety and without locking. For instance, both receivers in the \texttt{ImageCrop} component from Figure \ref{fig:CodeComponent} access the private \texttt{image} field: the image receiver writes to the field, whereas the rectangle receiver reads from it. However, because the runtime enforces component-level receiver exclusivity, no locking is actually required to access the image. Regardless of when the messages on the incoming streams arrive, the runtime makes sure that only one of these two receivers is executing at any given time, and in the process provides component-level state protection.

Receiver exclusivity goes hand-in-hand with a recommended design pattern that minimizes the amount of work done by each individual receiver. If an individual receiver does a lot of computation and takes a long time to execute, it can induce large latency in the pipeline, as it would prevent other receivers on the same component from being executed for long periods of time. The \psif programming paradigm encourages decomposition and encapsulation, and discourages large, monolithic components. Instead, the runtime provides the means for constructing larger components by hierarchically aggregating smaller ones, which we will review in more detail in Section \ref{subsec:Runtime_CompositeComponents}. The runtime not only fosters encapsulation and reuse, but also aims to maximize resource utilization and enable graceful scale-up via the parallel execution of individually lightweight components.

\subsubsection{Automatic Data Cloning}
\label{subsubsec:Runtime_Cloning}

Apart from receiver exclusivity, the \psif runtime implements a second important mechanism that insulates component developers from the complexities of concurrent execution and simplifies the component authoring process: \emph{automatic data cloning}. 

In Section \ref{subsubsec:Runtime_Scheduling}, we have described how each time a message is posted on an emitter, the runtime places \emph{a copy} of the message in the subscribing receiver's delivery queue. The copy is automatically generated by a cloning subsystem in the runtime, regardless of the type of the message, and creates isolation between components. When a message is posted to an emitter, each downstream component receives a separate copy of the message in their receiver methods, rather than a reference to the same message. As such, components are free to modify the contents of the incoming message inside receiver methods if needed, as these edits will not affect other components that may be processing the same message at the same time. Similarly, the automatic cloning mechanism allows component authors to post an object instance to an emitter, and then immediately modify the contents of that object instance: the outgoing message will not be affected, as the cloning mechanism in effect insulates the component from the rest of the computation graph. This pattern is particularly useful in cases where the results to be posted are maintained as saved members of the component's internal state, edited based on incoming messages and then posted to the emitter.

The cloning subsystem in the \psif runtime is automatic. The component authors can use any \textsc{.NET} data type, and do not need to write custom cloning code for any of the data types carried over streams; they simply create emitters and receivers of the desired types, and post data. The runtime automatically generates the necessary cloning code on the fly, by reflecting over the types involved and dynamically emitting code. The cloning subsystem creates \emph{deep} clones, and handles any \textsc{.NET} data types, including polymorphic objects and complex object graphs that contain self-references. A similar automatic subsystem is used to perform serialization when data is written to disk (see Section \ref{subsubsec:Runtime_Persistence}), or transported across remoting boundaries (see Section \ref{subsec:Runtime_RemotingInterop}). 

\subsubsection{Memory Management}

While automatic data cloning ensures component isolation, it comes at a potential performance cost in a managed memory framework like \textsc{.NET}: constructing new instances of every message passed on a stream can lead to increasing costs in garbage collection. Given the large number of messages flowing through streams, these costs could become significant. With large messages, such as images, additional costs can accrue due to the time involved in creating copies of the data. The \psif framework mitigates these costs by allowing developers to leverage the modern memory management affordances of \textsc{.NET}, while still retaining the runtime performance characteristics necessary for multimodal, integrative-AI applications. Specifically, the \psif runtime implements a set of memory management mechanisms that mitigate garbage collection costs and aim to achieve no new memory allocations once the pipeline reaches steady state. Data copying costs can also be completely avoided in some cases.

To minimize new object allocations and garbage collections, the \psif runtime associates a \emph{recycling pool} with each receiver, i.e., a buffer of message instances that provides storage for stream messages transported via the delivery queue into that receiver. Recycling pools are automatically constructed for any type of message. Once a receiver is finished processing an incoming message, that message is then automatically recycled back to the pool, and a subsequent incoming message can be cloned into the same memory location, without requiring a new allocation. When cloning an incoming message, new objects are therefore created only if the recycling pool has exhausted all available recycled object instances. This technique keeps object allocations to a minimum, and when the pipeline is running at a steady state, no new allocations are necessary. 

The recycling pool operations are automatic and completely transparent to component authors, with one important exception. Consider the case when an incoming message---or any portion of it---needs to be captured and held by the component for use beyond the point at which the receiver method finishes executing. In this case, the information from the message must be cloned locally, since the message itself will be released back to the recycling pool once the receiver method exits. The programming pattern for receiver methods is therefore similar to that of event handlers in \textsc{.NET}, where any data incoming via an event handler should not be directly used past the lifetime of the event handler, without first copying it. To enable easy copying of message information, the \psif runtime provides a \texttt{DeepClone()} method which allows component authors to use the same automatic cloning mechanism employed by the runtime to make their own copies of the incoming data.

Apart from minimizing new allocation via \emph{recycling pools}, the \psif runtime also provides mechanisms for avoiding memory copy costs for large, read-only objects such as images, depth maps, etc. This is accomplished via \emph{shared objects}, a mechanism that enables developers to declare certain objects as read-only and shared across components. For these messages, the automatic cloning mechanism simply copies the reference, rather than performing a deep copy of the instance. Components that receive messages with shared objects must respect the implicit contract that shared objects are not to be modified. Shared objects are also generally allocated via recycling pools corresponding to their type, and \psif implements a \emph{reference counting} mechanism that enables the runtime to reason about each shared object's lifetime. This mechanism allows the runtime to determine when a specific instance is no longer used anywhere in the pipeline and can be returned to its recycling pool. For more details about memory management, and the implementation and function of recycling pools and shared objects, the interested reader is referred to \cite{PsiWebsite_SharedObjects}.

\subsubsection{Persistence}
\label{subsubsec:Runtime_Persistence}

The \psif runtime provides serialization and persistence mechanisms that enable logging of streams to persisted data stores. The serialization mechanism is automatic and similar to the data cloning mechanism we have described above in Section \ref{subsubsec:Runtime_Cloning}. Component and application developers generally do not need to write custom serialization code for the data types they use. Instead, the runtime automatically generates efficient serialization code dynamically by reflecting over the stream types. The serialization mechanism can handle polymorphic types, as well as types that contain cyclic references. The runtime also provides a set of APIs that enable developers to write their own custom serialization and cloning code if desired.

The persistence mechanism enables writing and reading streams to and from disk or to volatile in-memory stores. Persistence is performed with memory-mapped files and is optimized for throughput: serialized messages on streams from a given application are written interleaved, as they arrive, within contiguous file extents of a specified size (by default 256MB). In addition, a catalog containing stream metadata information (e.g., stream start and end times, number of messages, average message latency, etc.) is also persisted. Finally, the persistence system can be configured such that certain specified streams that contain large data types, for instance images, are written to a separate set of file extents and augmented with indexing information that facilitates seeking by time.

The persistence mechanism also enables \emph{data replay} scenarios, where streams can be ``played back'' from a persisted store, which can be very useful when iteratively tuning an application or component. The runtime provides a special component, called an \emph{importer}, that can read and surface streams from disk. Developers can control which portions of a store to play back, as well as the speed at which playback occurs. Two playback speeds are currently supported: real time (replicating the speed of the source streams when they were live), or as fast as possible (i.e., as fast as the messages can be read from disk). For more information on data replay, we refer the interested reader to the documentation available in \cite{PsiWebsite_PipelineExecution}. Finally, the runtime provides an interface specification that allows developers to write their own importers that can surface streams from third party data stores \cite{PsiWebsite_ThirdPartyStreamReaders}, such as MPEG files, WAV files, etc.

\subsection{Stream Operators}
\label{subsec:Runtime_StreamOperators}

In the previous section we introduced the core mechanisms underlying the \psif runtime, and described how pipeline execution takes place. We now turn our attention to \emph{stream operators}, which are a core set of components that enable manipulating generic streams of data. Later on, in Section \ref{sec:Components} we discuss the broader component ecosystem in \psifnospace, which includes higher-level components that wrap specific sensor or processing technologies, such as cameras, microphones, speech recognizers, object detectors, etc.

\subsubsection{Basic Stream Operators}

\begin{figure}
    \centering
    \fbox{\includegraphics[width=0.65\textwidth]{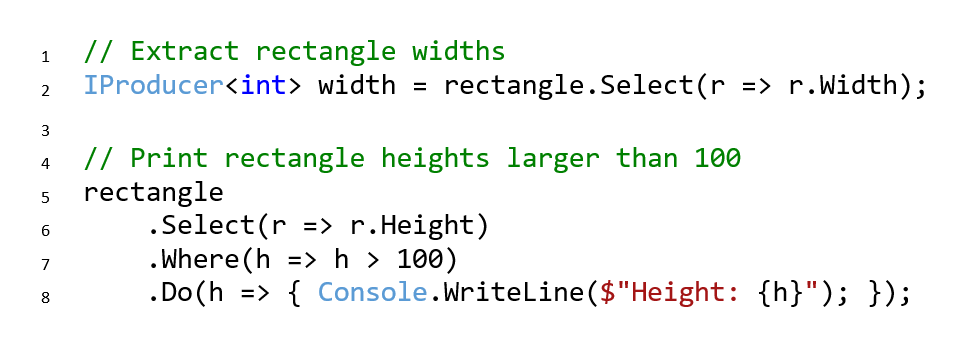}}
    \caption{Example code snippet showing basic \psif stream operators, including \texttt{Select()}, \texttt{Where()}, and \texttt{Do()}.}
    \label{fig:StreamOperators}
\end{figure}

\psif includes a set of basic stream operators for processing generic streams of data. One such operator is \texttt{Select()}, which allows developers to transform an input stream by applying a selector function to each message in the incoming input. Figure \ref{fig:StreamOperators} shows an example in line 2 of the code snippet: a stream of rectangles---a data structure with fields \texttt{Left}, \texttt{Top}, \texttt{Width}, and \texttt{Height} of type \texttt{int}---is transformed into a stream of \texttt{int} values containing the width of each rectangle on the input stream. The \texttt{Select()} operator takes as a parameter a lambda function that transforms every incoming message. In this case, for every rectangle \texttt{r}, it returns the corresponding width of \texttt{r}. The result of the \texttt{Select()} operator is the new stream, this time of type \texttt{int}, containing the rectangle widths.

The \texttt{Select()} stream operator illustrated above is backed by a \psif component that has a single receiver of a generic type \texttt{TInput} and a single emitter of a generic type \texttt{TOutput}. The stream operator itself is implemented as an extension method for generic streams of type \texttt{TInput}, i.e., \texttt{IProducer<TInput>}. The extension method constructs the underlying component, connects the specified input stream to the component receiver via the \texttt{PipeTo()} mechanism we have already seen, and returns the output stream produced by the component. 

The pattern described above, in which stream extension methods are used to create stream operators that encapsulate components, is common throughout the framework. The pattern enables developers to easily chain multiple computations together, as shown in Figure \ref{fig:StreamOperators} on lines 5--8. In this example, we print the incoming rectangle heights, but only when these heights are larger than 100. Line 6 uses a \texttt{Select()} operator to generate a stream with the height of each input rectangle. Line 7 uses a \texttt{Where()} operator over the resulting stream to only pass messages through which are above the specified height threshold. Finally on line 8, a \texttt{Do()} operator simply performs a specified action for each message---in this case printing them to the console.

In conjunction with \texttt{PipeTo()}, these basic operators facilitate connecting components, and allow developers to easily construct pipelines that resolve API mismatches between disparate components as needed, while keeping the streams strongly typed. For example, if a face detector outputs instances of type \texttt{Face}, but a downstream image crop component requires \texttt{Rectangle} instances, the application developer can easily connect these components via a \texttt{Select()} stream operator that transforms the stream of faces into a stream of rectangles.

Besides \texttt{Select()}, \texttt{Where()}, and \texttt{Do()}, the \psif framework provides other basic operators that simplify processing of generic streams of data: these include a number of time-related operators, mathematical and statistical operators, aggregating operators, and windowing operators. Another category of basic operators are \textit{stream generators}, which can produce streams of data based on an existing pattern, e.g., by repeating a specified value, applying a function to incrementally compute the next value, or outputting messages from an existing enumeration. A full description of all operators available in the framework is beyond the scope of this paper; we refer the interested reader to the documentation available in \cite{PsiWebsite_BasicStreamOperators}. In the next two sub-sections, we do however describe in more detail two additional important sets of operators: those that enable stream fusion and merging, and those that enable dynamic pipelines.

\subsubsection{Stream Fusion and Merging Operators}

Multimodal applications often need to fuse data arriving with different latencies on different streams. For instance, in the example we introduced earlier in Figure \ref{fig:SimpleAppPipeline}, the multimodal voice activity detector fuses information arriving on two different streams, and derived from two different channels: audio and video. The \psif runtime provides a number of stream operators that facilitate both \emph{stream fusion} and \emph{stream merging}, and provide the basis for performing these operations in a correct and reproducible manner.

\begin{figure}
    \centering
    \includegraphics[width=0.65\textwidth]{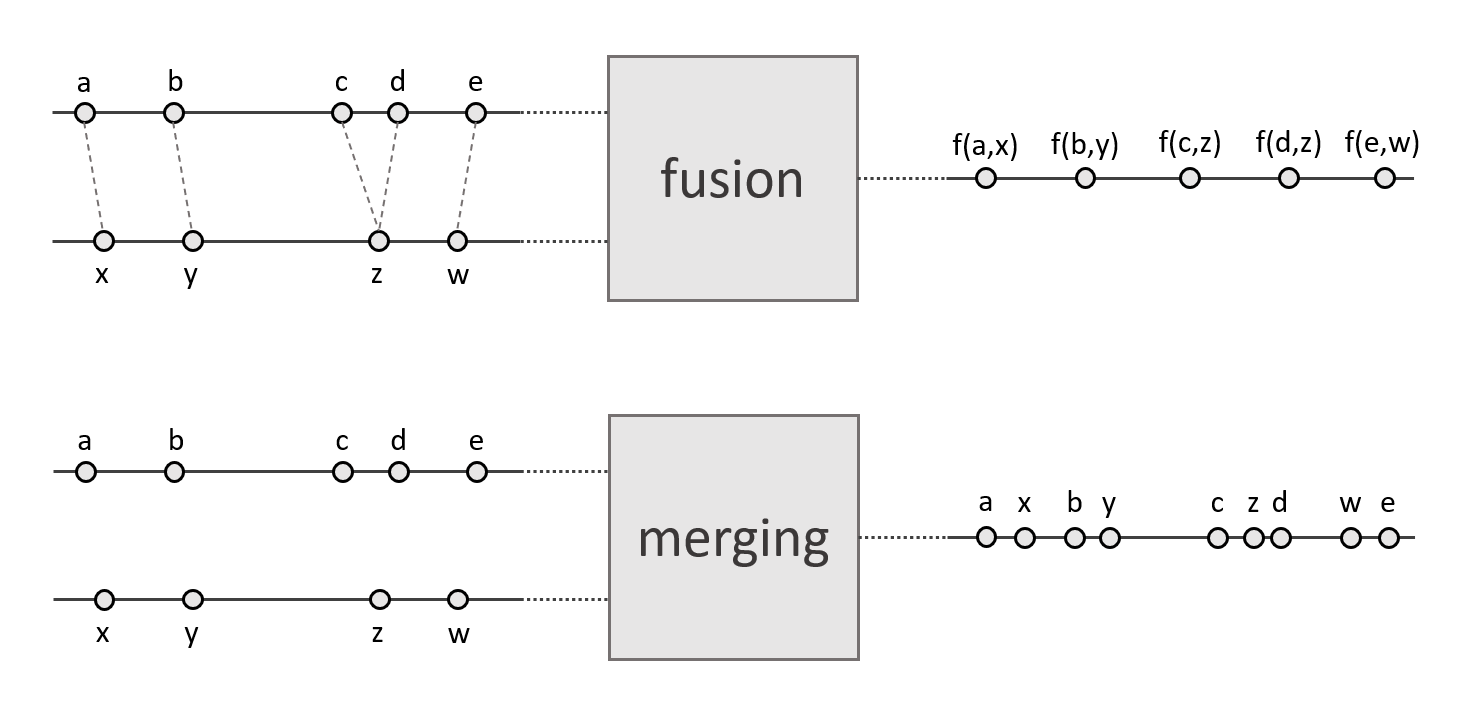}
    \caption{Top: In stream fusion, an association is formed between messages across the input streams, and output messages are created by applying a function to each set of associated incoming messages. Bottom: In stream merging, input messages from multiple streams are emitted in a particular order on an output stream.}
    \label{fig:FusionAndMerging}
\end{figure}

In \emph{stream fusion} operations, messages from two or more streams are fused together to construct new output messages. Each output message is created by applying a function to sets of messages selected from the incoming streams, as shown in Figure \ref{fig:FusionAndMerging}, top. The fundamental problem that has to be resolved is how to associate the messages on the input streams, taking into account the fact that these messages may arrive at different latencies.

\begin{figure}
    \centering
    \includegraphics[width=0.35\textwidth]{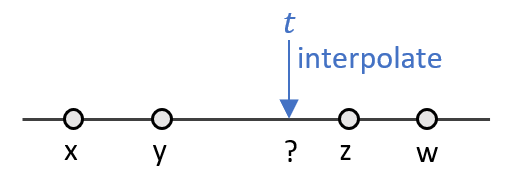}
    \caption{A stream interpolator function constructs an interpolation result for time \texttt{t} in a stream containing messages \texttt{x}, \texttt{y}, \texttt{z}, and \texttt{w}.}
    \label{fig:Interpolator}
\end{figure}

\begin{figure}
    \centering
    \includegraphics[width=\textwidth]{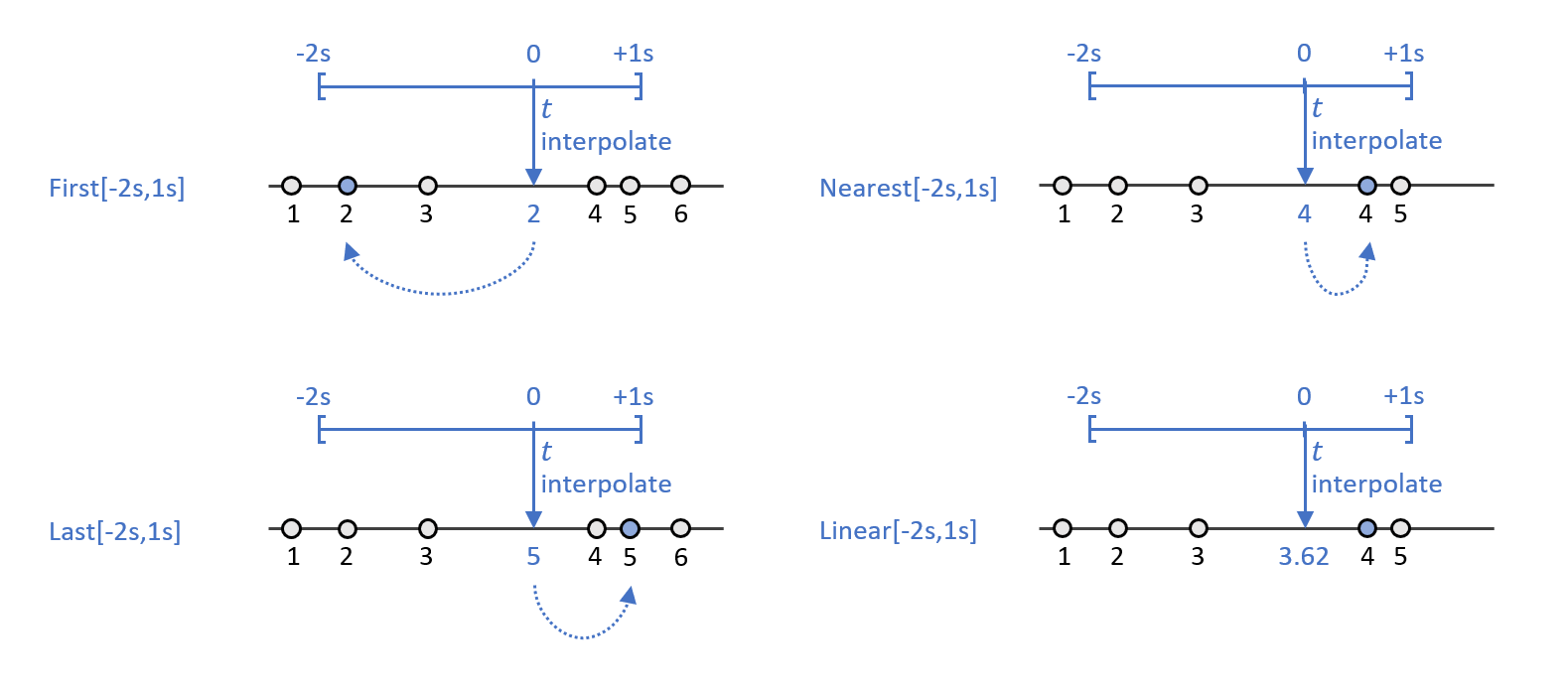}
    \caption{Example window-based interpolation operations and results. For a window of time stretching 2 seconds backward and 1 second forward of each primary stream message (``0''), examples are shown for selecting from the secondary stream the first (``2''), nearest (``4''), or last (``5'') message in the specified window, or computing a linear interpolation (``3.62'') at the specified time.}
    \label{fig:WindowBasedInterpolators}
\end{figure}

The \psif framework resolves this problem by providing a general \emph{interpolator} mechanism. Specifically, an \emph{interpolator} is a function that constructs an interpolation result corresponding to a specified interpolation time in a stream of messages (see Figure \ref{fig:Interpolator}). \psif provides a variety of interpolators that simplify data fusion operations. Interpolators can either sample---i.e., select---an existing message from the stream, or create a new message altogether. For example, a set of \emph{window-based message-sampling interpolators} enable selecting the first, last, or nearest message in a specified time window relative to the interpolation time, as illustrated in Figure \ref{fig:WindowBasedInterpolators}. In contrast, a linear interpolator can create new values by performing linear interpolation on a numerical stream, as illustrated on the bottom, right-hand side of Figure \ref{fig:WindowBasedInterpolators}. Developers can also write new stream interpolators to implement other types of interpolation or sampling mechanisms. For a full description of the set of available interpolators, we refer the interested reader to \cite{PsiWebsite_StreamFusionAndMerging}.

To perform stream fusion, \psif provides a \texttt{Fuse} component and corresponding stream operator parameterized with an interpolator. This component takes in a primary stream and one or more secondary streams. It buffers the inputs on these streams as necessary, and for each input it receives on the primary stream, it calls the interpolator to compute the corresponding messages on the secondary streams. It then applies an optional function to these messages to construct a fused output, or else simply emits them in a tuple.

An important aspect of the stream fusion component and available set of interpolators is that they enable \emph{reproducible fusion}. As we have observed before, the messages on the incoming streams may arrive at different latencies. If the fusion component would pair the incoming messages as they arrive, the results would depend on the latency of the input messages, and would not be reproducible, i.e., performing fusion with the same source streams again at a later time or in a different execution environment might lead to different results due to differences in times of arrival.

To enable reproducible fusion, the framework provides a set of reproducible interpolators which reason about the originating-times of messages. When using these interpolators, the fusion component waits to emit an interpolation result until it can guarantee that the result is correct and reproducible, regardless of the latencies of the incoming messages. For instance, the \texttt{Reproducible.Nearest(window)} interpolator determines the nearest message on the secondary stream to the originating-time of a given primary stream message, within a specified window. Using this interpolator, for each incoming message on the primary stream, the \texttt{Fuse} component waits until a secondary stream message arrives that ensures that the computed interpolation result is indeed the correct one. Consider the concrete example shown on the top, right-hand side of Figure \ref{fig:WindowBasedInterpolators}, i.e., \texttt{Nearest[-2s, 1s]}. In this case, for the primary stream message indicated by ``0'', the \texttt{Fuse} component has to wait until message ``4'' arrives on the secondary stream before it can emit a result. Simply seeing messages ``2'' and ``3'' is not sufficient, even though they are within the specified window, because the next message that arrives might be closer in time to ``0'' than ``3'' is---which indeed is the case with message ``4'' in this example. Once message ``4'' arrives, the component knows that no subsequent message can be nearer to the primary message (since originating-times of messages must be strictly increasing on each stream), therefore it can emit a result. In general, the messages arriving on the secondary stream can have arbitrary latency, and the \texttt{Fuse} component does not generate an output corresponding to each primary stream message until it is certain that output is correct, i.e., no future messages arriving on the secondary stream can change this result.

When using reproducible interpolators, fusion results only depend on the originating-times and contents of the input messages, and not on any computational delays. However, this behavior might induce additional delays as the fusion component waits to produce outputs until results are verifiably correct. The framework also provides a set of interpolators that do not wait for verifiably correct inputs based on originating-times, but rather provide the first, last, or nearest message available in the relative window at the moment that the primary input arrives, e.g., \texttt{Available.Nearest(window)}. Results from these interpolators are necessarily not reproducible.

In contrast to stream fusion, in a \emph{stream merging} operation, the elements arriving on multiple input streams are merged in a specific order on the output stream, as illustrated in Figure \ref{fig:FusionAndMerging}, bottom. Like with stream fusion, \psif provides an operator, called \texttt{Zip()}, that enables reproducible merging based on originating-times. The corresponding \texttt{Zip} component buffers inputs as necessary and merges the messages arriving on multiple streams in the order of their originating-time. Like with fusion, because messages on different streams may arrive at different latencies, \texttt{Zip} may have to wait for future messages to arrive and as such can induce extra delay to generate a reproducible result. The framework provides a fast \texttt{Merge()} operator as well, that sacrifices reproducibility and posts the incoming messages on the output stream immediately, regardless of originating-time.

Additional information and a more in-depth discussion of stream fusion and merging operators is available for the interested reader in the online documentation \cite{PsiWebsite_StreamFusionAndMerging}. Both with fusion and merging, \psif enables reproducibility of results, which is a core, important property and benefit of the \psif streaming infrastructure. In conjunction with the replay capabilities of the runtime, this reproducibility enables developers to more reliably debug and perform iterative tuning of applications and their components.

\subsubsection{Dynamic Pipelines and the Parallel Operator}
\label{subsec:Runtime_DynamicPipelines}

In the examples discussed so far, \psif pipelines were constructed statically via imperative code written by the application programmer. Once the \texttt{Run()} method is invoked, the runtime is in charge of pipeline execution, and the structure of the pipeline is fixed. The \psif runtime however also enables developers to construct pipelines that can dynamically change their structure throughout execution, depending on the data flowing through the streams.

\begin{figure}[!b]
    \centering
    \includegraphics[width=\textwidth]{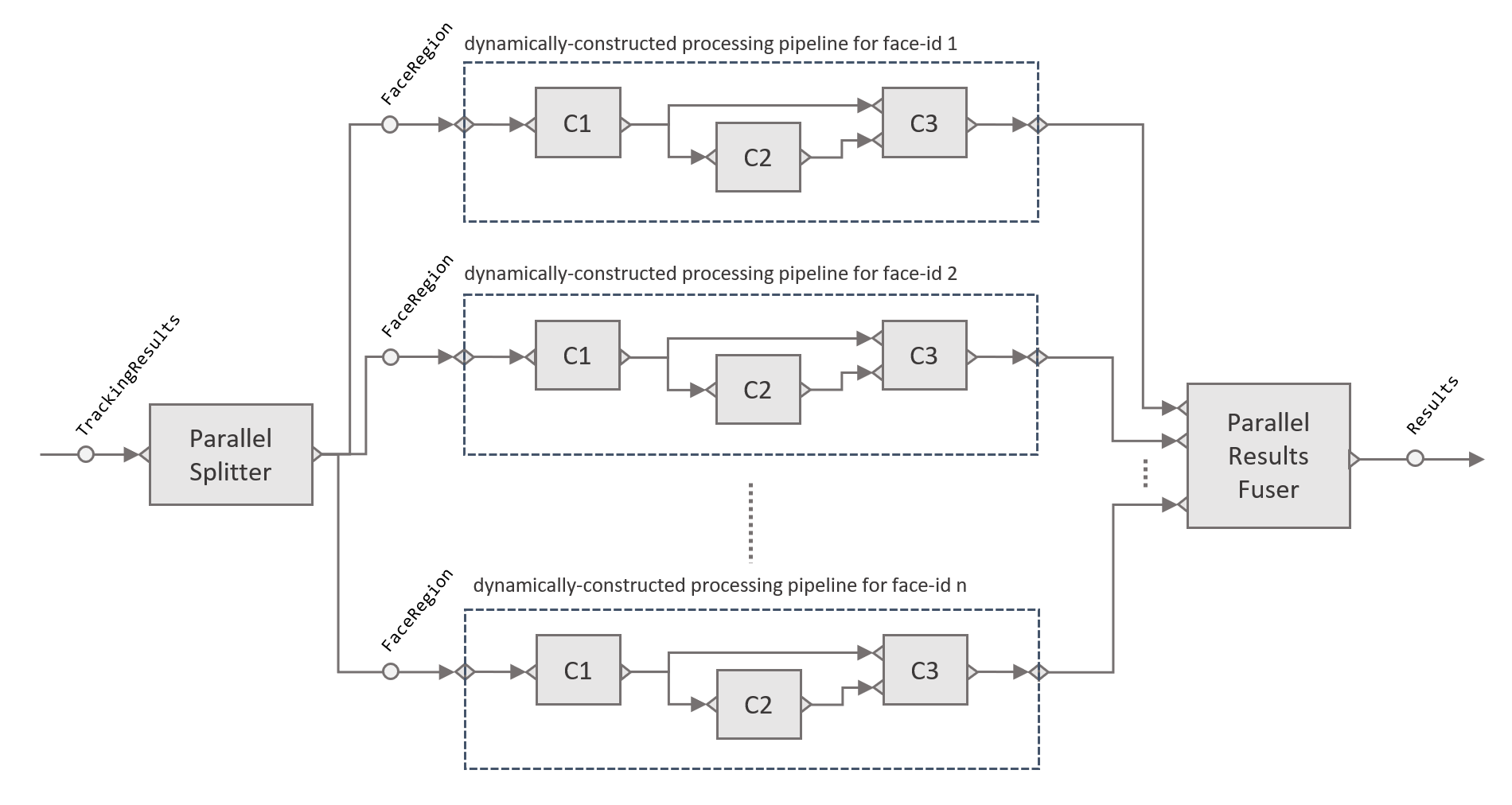}
    \caption{Illustration of dynamic pipelines constructed via the \texttt{Parallel()} operator. Lists of face tracking results are split into separate streams of \texttt{FaceRegion} based on face-id. A subpipeline containing components C1, C2, and C3 is dynamically constructed for each stream corresponding to a unique face-id. The results of each dynamic subpipeline are then fused together.}
    \label{fig:ParallelOperator}
\end{figure}

A common scenario for the use of dynamic pipelines is one in which we would like to instantiate the same processing pipeline multiple times for each instance of a particular input object. For example, consider an application that performs face tracking over images generated by a camera. In this scenario, we may wish to dynamically instantiate and execute a processing \emph{subpipeline} for each newly detected face, for instance to perform additional analyses such as affect recognition on the face region. This subpipeline might consist of multiple chained components, each of which may maintain internal state. A new subpipeline would need to be dynamically created each time a new face is detected, and it will also need to be shut down each time a face disappears. Multiple subpipelines may be running in parallel if multiple faces are simultaneously present, and results from these pipelines should be merged seamlessly into a single stream.

The \psif runtime enables this pattern of dynamic pipelines via a specialized stream operator, called \texttt{Parallel()}. In the most general version, this operator is parameterized by a \emph{key-generator function} that generates key-value pairs from the messages in the incoming stream, and a \emph{pipeline-generator function} that generates a subpipeline for each unique key and associated stream of values. For instance, in the example above, the key generator function may transform the incoming list of tracked faces into pairs of (face-id, face-region), and the pipeline-generating function specifies how to construct a subpipeline that processes a stream of face-regions corresponding to a face-id. For each new key (face-id) that appears, the pipeline-generator function is invoked and constructs a new subpipeline. These subpipelines do not necessarily need to be identical---the computation graph instantiated for each key could have a different structure. 

The results of the individual subpipelines instantiated by the \texttt{Parallel()} operator for each face-id are fused back into a single stream. The overall structure of this dynamic pipeline is illustrated in Figure \ref{fig:ParallelOperator}. An in-depth discussion of the full capabilities of this operator is beyond the scope of this overview paper---we refer the interested reader to the online documentation available in \cite{PsiWebsite_ParallelOperator}.

\subsection{Composite Components}
\label{subsec:Runtime_CompositeComponents}

In the previous section we have reviewed the array of primitive operators that provide the basis for processing, fusing, and merging streams of data, as well as for constructing dynamic pipelines. Basic components and stream operators are generally lightweight---e.g., applying a function on each message, filtering messages, sampling, interpolating, etc. The APIs and overall computational model set forth in the \psif framework promote a programming style where components are lightweight and can be easily chained together to accomplish larger computation tasks. Because components can execute concurrently, this programming style facilitates efficiency gains via pipeline parallelism.

The lightweight nature of \psif components also means that \psif applications can quickly grow large in terms of the number of components involved. To address the challenges that arise with developing and maintaining pipelines of increasing size and complexity, \psif allows for organizing components and pipelines hierarchically. Specifically, a graph containing multiple components can be encapsulated and used as a single \emph{composite component}. To the application developer and to the outside pipeline, this graph of components appears as if it was a single component. The runtime, however, still reasons about and performs scheduling over the individual, atomic components, retaining the performance benefits of fine-grained pipeline parallelism.

\begin{figure}[!t]
    \centering
    \includegraphics[width=\textwidth]{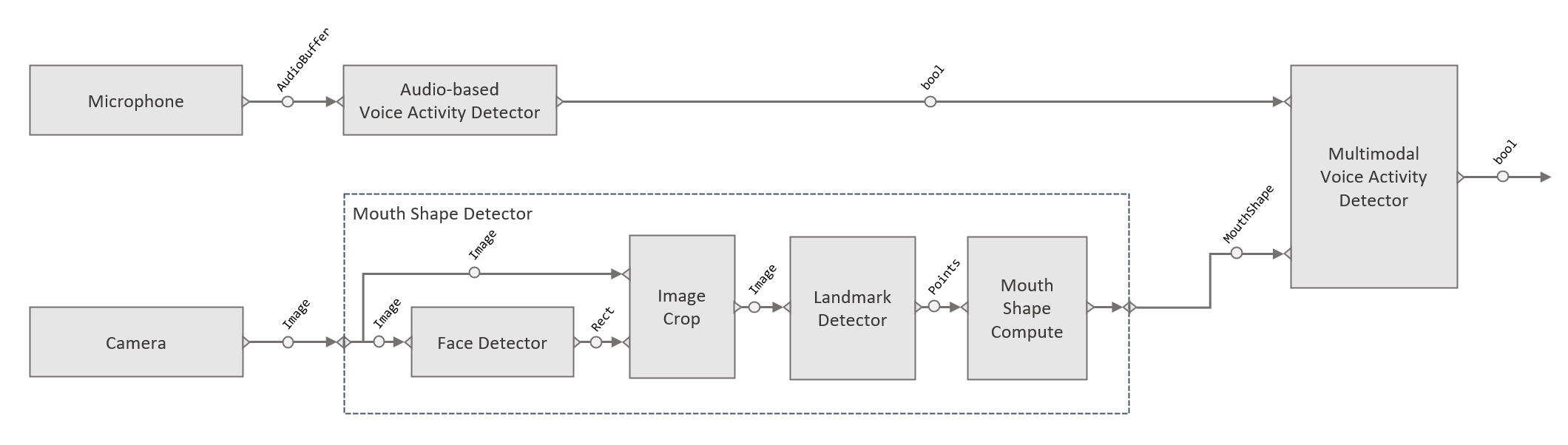}
    \caption{Illustration of the Mouth Shape Detector as a composite component, comprising a graph of sub-components for face detection, image cropping, landmark detection, and finally mouth shape computation. The application developer can instantiate and use this graph as if it were a single component, but the \psif runtime performs scheduling over the atomic components underneath.}
    \label{fig:CompositeMouthShapeDetector}
\end{figure}

For instance, in the multimodal voice activity detection example illustrated in Figure \ref{fig:SimpleAppPipeline}, the \texttt{MouthShapeDetector} component could in fact be implemented as a \emph{composite component}. The code shown in Figure \ref{fig:SimpleAppCode} would stay the same: the component can be instantiated and connected in the application pipeline like any other component, and it still exposes a receiver of images and an emitter of mouth shape detection results. However, internally, this component could be composed of multiple other components, as illustrated in Figure  \ref{fig:CompositeMouthShapeDetector}: the input image stream is passed both to a face detector, and to an image-crop component which uses this stream in conjunction with the face detection results to crop a smaller image around the face. This image is passed to a landmark detector, and the resulting aligned face points are used to compute the mouth shape.

The \psif framework provides APIs for easily constructing such composite components, and we refer the interested reader to the online documentation \cite{PsiWebsite_WritingComponents} for the technical details. We note that the hierarchical compositing of components is not limited to just one level. The \texttt{MouthShapeDetector} component could itself be part of a larger composite component, and so on. Coupled with advanced tools for visualizing the structure of the pipelines, which we will discuss in Section \ref{subsec:Tools_PipelineVisualization}, the ability to organize components hierarchically enables encapsulation and simplifies development efforts, all while retaining the efficiency gains from fine-grained parallelism.

\subsection{Remoting and Interop}
\label{subsec:Runtime_RemotingInterop}

As we have seen, a typical \psif application is implemented as a pipeline of components connected via streams of data. The application executes in a single process and the \psif runtime is in charge of delivering messages over the streaming connections and orchestrating the execution of the component receivers. However, \psif also allows for developing distributed applications composed of multiple coordinated processes, each running a separate pipeline, through a mechanism called \emph{remoting}. Through this mechanism, data are serialized and transmitted between processes and/or machines over TCP, UDP or Named Pipes.

The remoting connection can be implemented with the help of a \emph{remote exporter} and a \emph{remote importer} component. The former, implemented by the \texttt{RemoteExporter} class, presents a similar API to a regular store exporter which is used to persist streams for logging purposes. Similarly, on the receiving side, a \texttt{RemoteImporter} class presents a similar API to a regular store importer which is used to read and replay streams from a persisted store. In fact, the remoting system is based on local stores on each end. These may be volatile, in-memory only stores, or may be persisted. One advantage of backing the communication with actual data stores is fault tolerance. If the connection is broken, messages continue to be written locally and will be relayed upon re-connection.

The transmission protocol always uses TCP for carrying stream catalog information (i.e., stream metadata), but may use Named Pipes, UDP or TCP for the actual message data. The message and stream catalog payloads are generated by the same serialization system, and are therefore byte-for-byte identical to the persistence format used by \psif stores. Message delivery over TCP or Named Pipes is guaranteed. With UPD, individual packets may be dropped, delivered multiple times or delivered out of order, but the remoting protocol guarantees ordering and packet reassembly into atomic messages; however entire messages may be dropped. If the network cannot keep up with messages being written to a remote exporter, then throttling backpressure will propagate up the pipeline and messages will be handled according to the specified delivery policies. An arbitrary bytes-per-second quota and averaging window may also be configured on the remote exporter to propagate backpressure when the average threshold is exceeded.

For integration with ecosystems and languages outside of \textsc{.NET}, \psif provides general \emph{interop} capabilities and APIs. These capabilities may be useful to, for example, interop with Python code, or with JavaScript for web dashboards and Node.js, and so on. In general, interop in \psif is accomplished by converting \psif data messages into standard formats such as JSON (text), MessagePack\footnote{\url{https://msgpack.org/}} (binary), and comma-separated values (non-hierarchical). All data may then be persisted to flat files or conveyed over standard transports such as ZeroMQ\footnote{\url{https://zeromq.org/}}. Additional custom formats and transports can be implemented by developers through the use of the provided serialization and transport interfaces.

For the special case of integration with ROS (Robot Operating System) \cite{quigley2009ros}, \psif specifically provides a \emph{ROS bridge} which may be used to integrate ROS components into a \psif system. A ROS system can be integrated with a \psif application by constructing components utilizing the bridge to communicate with the ROS Master, negotiate peer-to-peer connections, and become proper publishers and subscribers over ROS messaging protocol, which are exposed as emitters and receivers in the \psif world. We refer the interested reader to the samples and documentation available in \cite{PsiWebsite_ROSIntegration}.

\section{Tools}
\label{sec:Tools}

Multimodal, integrative-AI applications are characterized by a number of aspects that significantly increase challenges with debugging, testing and development. These applications often leverage multiple components that execute concurrently yet need to be coordinated with each other. They manipulate a wide array of data types, such as audio, video, depth, numerical signals, etc., and the development cycle often includes data-driven analyses and iterative tuning of various components. 

\begin{figure}[!t]
    \centering
    \includegraphics[width=\columnwidth]{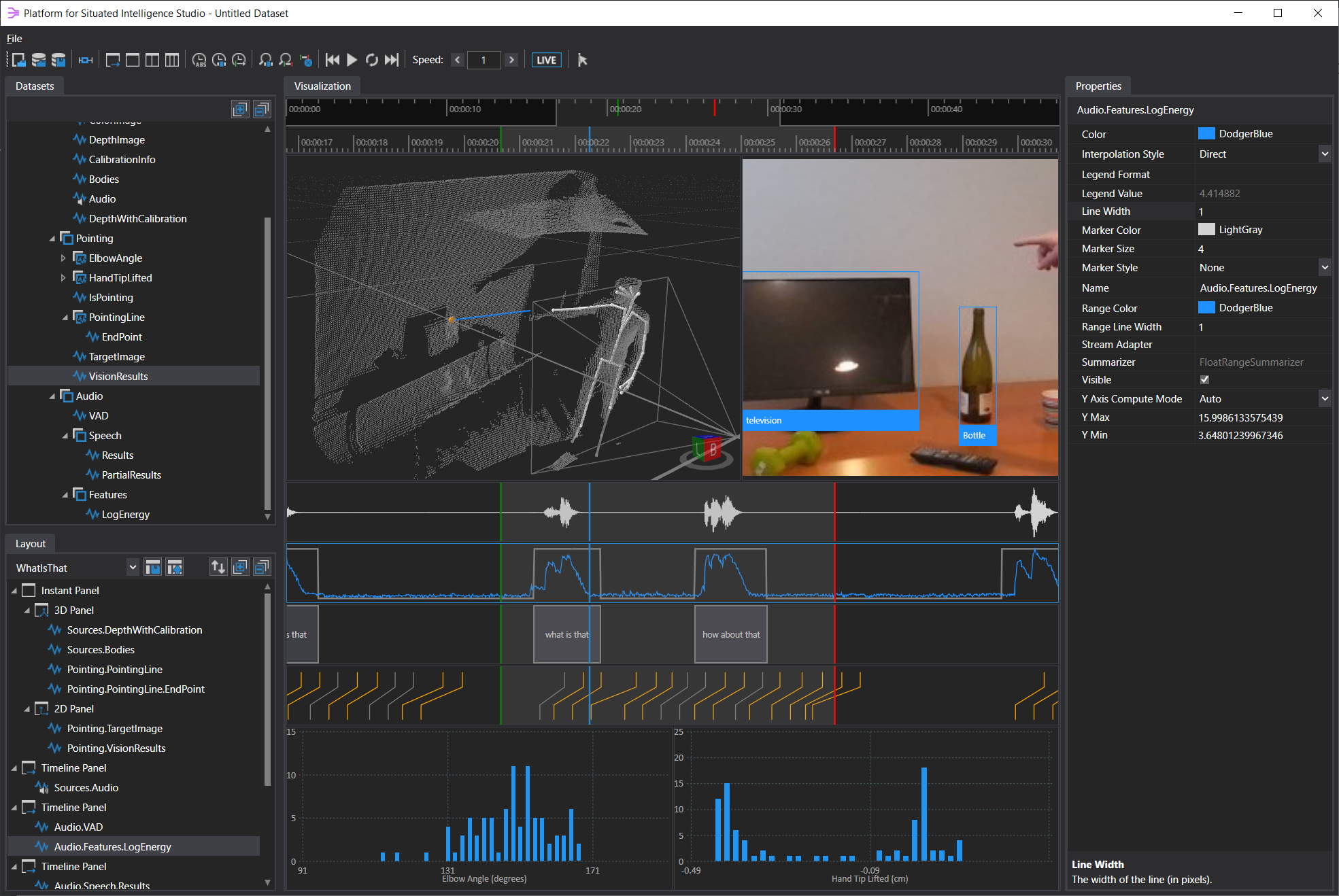}
    \caption{Platform for Situated Intelligence Studio: a tool for multimodal data visualization, annotation and processing.}
    \label{fig:PsiStudio-Main}
\end{figure}

Given the concurrent yet coordinated nature of these applications, standard debugging techniques such as the use of breakpoints or ad-hoc logging of debug information (a.k.a. ``printf debugging'') are not efficient in these settings. The ability to visualize the data flowing through the application over time, and perform analyses and queries over it becomes paramount and can significantly accelerate the development process.

To this end, \psif provides a set of tools and APIs that enable multimodal data visualization, annotation, and processing to accelerate application debugging, maintenance, and tuning. \emph{Platform for Situated Intelligence Studio}, or in short \psistudionospace, is a GUI-based data visualization and analytics tool, shown in Figure \ref{fig:PsiStudio-Main}. \psistudio allows users to visualize the various data streams persisted by a \psif application, as well as the structure of the application pipeline, together with pipeline performance and diagnostic information. These visualizations can be performed both \emph{offline}, i.e., based on data stores persisted by the application, and also \emph{live}, while the application is running.

A wide array of \emph{visualizers} for different data types, from simple numerical streams, to audio, video, complex 3D objects, etc., are available. The visualizers have configurable properties, and can be composited into complex \emph{visualization layouts} which can be persisted and reused. Users can easily navigate temporally through the data, inspect values, and select and play-back segments at varying speeds. In addition to basic visualization tasks, \psistudio also supports \emph{batch data processing}, as well as \emph{data annotation} scenarios for multimodal applications.

A sibling command-line tool, called \psistorenospace, is also available and enables batch data processing and analysis. In the subsections that follow, we review these tools and their affordances and associated APIs in more detail. We begin by discussing the capabilities for visualizing streaming data in Section \ref{subsec:Tools_DataVisualization} and the pipeline structure in Section \ref{subsec:Tools_PipelineVisualization}, followed by a discussion of data processing and data annotation in Sections \ref{subsec:Tools_DataProcessing} and \ref{subsec:Tools_DataAnnotation}.

\subsection{Data Visualization}
\label{subsec:Tools_DataVisualization}

\psistudio enables visualization of temporal, multimodal data streams persisted in \psif stores (as well as in other streaming file formats.) The user interface is divided into four major areas, as illustrated in Figure \ref{fig:PsiStudio-Main}. On the left-hand side, the \textsl{Datasets} tab shows the streams available in the currently opened store. Below it, the \textsl{Layout} tab shows the various visualizers that have currently been instantiated. The \textsl{Properties} tab on the right-hand side displays properties of the streams such as number of messages, average message size, average throughput and latency, etc., or properties of the visualizers, such as line style and color. Finally, the visualization canvas area in the center of the user interface contains a configurable set of \emph{visualization panels} which in turn contains the stream \emph{visualizers}.

The \textsl{Datasets} tab displays the streams contained in the currently opened store. \psistudio organizes the streams hierarchically by treating the dot (\texttt{.}) characters in a stream name as delimiters. For instance, streams named \textsl{Sources.ColorImage}, \textsl{Sources.DepthImage}, \textsl{Sources.CalibrationInfo}, and so on, are all shown as \textsl{ColorImage}, \textsl{DepthImage}, and \textsl{CalibrationInfo} respectively, under a parent node named \textsl{Sources}. This convention, and the structure it induces, allows the developers to organize streams hierarchically, simplifying navigation and access when working with stores containing a large number of streams.
 
Streams can be visualized by directly dragging and dropping them from the \textsl{Datasets} tab onto an existing visualization panel, or onto the empty area in the visualization canvas, in which case a new panel is created to hold the stream visualization. Multiple types of visualizers may be available for a given stream---right-clicking on a stream opens up a context menu that provides access to all the applicable visualizers.

The set of visualization panels and visualizers are also reflected in the \textsl{Layout} tab, on the bottom-left. Clicking on a visualizer in this tab allows the developer to access, inspect, and modify various aspects of the visualizer (via the \textsl{Properties} tab on the right-hand side), such as the line color for a plot, the sparsity of the point cloud, etc. The current structure and configuration of the visualization panels and visualizers instantiated constitute a \emph{layout}. The user can construct, save, and easily switch between multiple such layouts; the current layout is displayed in a drop-down combo-box at the top of the \textsl{Layouts} tab.

Finally, above the visualizations canvas area, \psistudio displays a \emph{time navigator}, which shows the timeline for the store and the current view. The navigator provides timing information and facilitates temporal browsing and playback.

\subsubsection{Visualization Panels and Visualizers} \label{sec:Visualizers}

Visualization panels act as containers for stream visualizers, and fall into two categories: \emph{timeline panels} and \emph{instant panels}. A \emph{timeline panel} can contain \emph{timeline visualizers}, i.e., stream visualizers that display information over time, such as a plot visualizer showing the values of a stream of doubles, or an audio visualizer showing an audio waveform. In contrast, \emph{instant panels} can contain one or more 2D or 3D panels with corresponding \emph{instant visualizers}, such as image or point cloud visualizers.

For instance, Figure \ref{fig:PsiStudio-Main} shows data from an application where a system attempts to identify the objects that a user is pointing to. The top row, on the left-hand side contains a 3D visualization panel, in which information from the depth map, body tracking, pointing direction, and pointing target streams is simultaneously displayed. The right-hand side of the top row also contains a 2D panel, which visualizes the stream containing the cropped image around the pointing target, as well as the overlaid visualizer showing the object detection results.

Below, there are four timeline panels: the first one shows the audio signal, the second shows the log-energy in the audio signal and voice activity detection results, the third shows speech recognition results, and the last shows a latency visualizer that provides information about the originating-time and latency of each object detection result (this visualizer was configured to show latencies larger than 500ms in orange). Finally, in the bottom row another instant panel contains two 2D-panels with histogram visualizers.

The example above provides a sampling from the set of visualizers available in the framework. The currently available set includes temporal visualizers for basic numerical types, as well as typical media streams: audio, video, depth, skeletal tracking, etc. A couple of universal visualizers (that operate over streams of any type) allow for visualizing the latency or originating-times of messages. Finally, several specialized visualizers, e.g., for 3D coordinate systems, 2D rectangles, etc., are included. The set of existing visualizers provides only a starting point. Developers can write their own visualizers, or \emph{visualization adapters}---which convert an existing type to a type handled by one of the existing visualizers---and can configure \psistudio to load these 3rd-party visualizers and adapters from external assemblies \cite{PsiWebsite_ThirdPartyVisualizers}.

Visualizers typically expose several specific properties, for instance the color and transparency for the depth map, the line color for a plot, the bone diameter for a 3D tracked body skeleton, etc. These attributes can be actuated from the \textsl{Properties} tab. Visualization panels can be resized and reordered, and multiple visualizers and panels can be composited into complex layouts, which can be persisted to disk and reused. The tool facilities quickly switching between existing layouts, by simply changing a selection in a drop-down combo-box in the \textsl{Layout} tab.

\subsubsection{Temporal Navigation}

\psistudio enables easy temporal navigation through the data. Moving the mouse over the time navigator or any timeline panel actuates a temporal \emph{cursor} (the vertical blue line in Figure \ref{fig:PsiStudio-Main}), and the instant visualizers update to reflect the value of the data, e.g., the depth map, RGB image, etc., at that point in time. An optional \emph{snap-to-stream} mode enforces that the cursor only snaps to the originating times of the messages in a specified stream. Zoom and pan operations are also supported, and the navigator bar at the top provides detailed timing information, both in absolute and relative terms. 

The tool allows users to easily create \emph{temporal selections}, which appear highlighted and are bordered by a green and red vertical line, denoting their start and end time respectively, as shown in Figure \ref{fig:PsiStudio-Main}. The temporal selection can be used to create data annotations, which we discuss later in Section \ref{subsec:Tools_DataAnnotation}, or to specify a data region for playback. A \textsl{Play} button on the toolbar enables \emph{data playback}, where the cursor is automatically advanced over the current selection, either in real-time, or at a specified speed, through the data.

\subsubsection{Live Visualization}

Apart from visualizing streaming data previously persisted by a \psif application, \psistudio also enables \emph{live visualization}, i.e., it allows users to visualize the data streams from a store that is actively being written to by a live, running \psif application. 

When opening a store, \psistudio automatically detects if this store is actively being written to, and if so, enters a \emph{live-mode}, signaled by the blue \textsl{Live} indicator on the toolbar. All the  visualization capabilities available for offline data are available in live-mode, with the only difference that in this case the temporal cursor does not follow the mouse, but rather automatically advances to the latest incoming message; temporal zoom capabilities are still available. In essence, the user can observe a live view of the data flowing through the application. 

The user can choose to drop out of live-mode by clicking the \textsl{Live} button in the toolbar, and in doing so regains control of navigation, the temporal cursor, creating selections, and even playback. This can allow a developer to inspect a particular moment in time, while the live application continues to run. The user can also toggle back into live-mode: the cursor catches up with the latest message and continues again to automatically advance as more data flows in.

\subsection{Pipeline Structure and Diagnostics Visualization}
\label{subsec:Tools_PipelineVisualization}

In the previous section we have seen how \psistudio enables multimodal temporal data visualization. In addition to visualizing streams of data, \psistudio also enables developers to visualize and inspect various aspects of the structure of the application pipeline, together with diagnostic information, which can further accelerate debugging.

\begin{figure}[!b]
    \centering
    \includegraphics[width=\columnwidth]{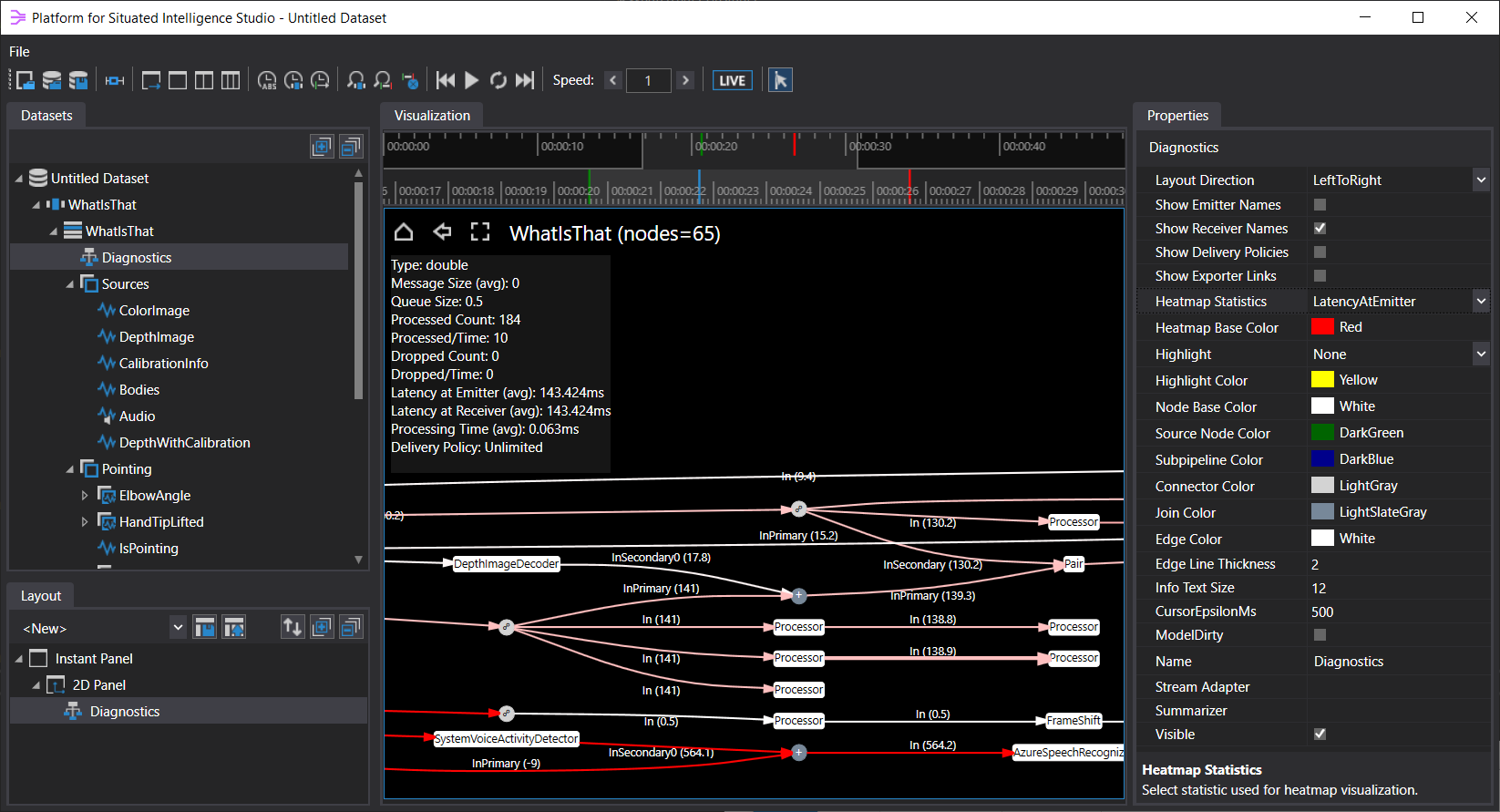}
    \caption{Platform for Situated Intelligence Studio: visualizing pipeline structure and diagnostics information.}
    \label{fig:PsiStudio-Diagnostics}
\end{figure}

To enable pipeline visualization, the application developer sets the \texttt{enableDiagnostics} parameter to \texttt{true} when creating the pipeline in the application code \cite{PsiWebsite_PipelineDiagnostics}. As a result, the pipeline exposes a stream, called \texttt{Diagnostics}, which can be persisted like any other stream, and visualized with a dedicated visualizer for pipeline structure. Figure \ref{fig:PsiStudio-Diagnostics} shows an example visualization. 

Using this visualizer, the developer can inspect the structure of the pipeline, including the various components and streams, and drill down hierarchically into subpipelines. The runtime collects and publishes on the diagnostics stream a number of important statistics about the pipeline execution, such as delivery queue sizes, number of messages processed and dropped on every stream, average message latency at emitters and receivers, etc. The diagnostics visualizer enables developers to easily access this information. For instance, clicking on a stream or a component shows various associated statistics; in addition, the visualizer can perform a heatmap coloring of the graph based on specified statistics. For instance, using a heatmap coloring based on average message latency enables the developer to quickly spot which components in a pipeline are slow; or using a heatmap coloring based on average throughput can help identify blockages and data synchronization problems. For more information, we direct the interested reader to the online documentation \cite{PsiWebsite_PipelineDiagnostics}.

\subsection{Data Processing}
\label{subsec:Tools_DataProcessing}

In general, each run of an application will generate a data store. The \psif infrastructure, as well as \psistudionospace, support working with multiple stores by leveraging a \emph{dataset} construct. A \emph{dataset} consists of multiple \emph{sessions}, where a session represents a temporal segment of data collected from a run of an application. Each session in turn can subsume one or more \emph{partitions}, where each partition is an actual data store. Partitions enable scenarios where a single application might simultaneously write to multiple stores (partitions), which can be grouped together and viewed as a single session. Figure \ref{fig:Dataset} below shows the overall structure of an example \psif dataset.

\begin{figure}[!t]
    \centering
    \includegraphics[width=\columnwidth]{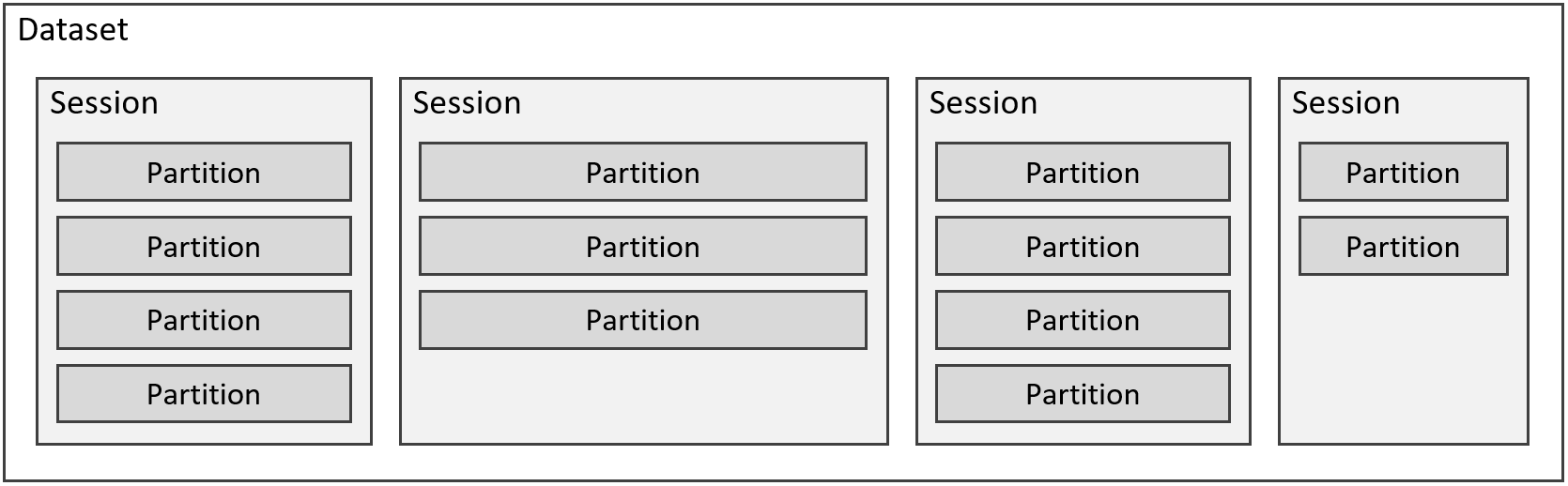}
    \caption{Structure of an example \psif dataset consisting of multiple sessions, each representing a temporal segment of persisted data. Each session subsumes one or more partitions, each of which corresponds to an actual data store.}
    \label{fig:Dataset}
\end{figure}

Each time a store is opened in \psistudionospace, the tool automatically creates a dataset around it: the dataset contains a single session, and that session contains a single partition (i.e., the specified store), as shown in the \textsl{Datasets} tab from Figure \ref{fig:PsiStudio-Main}. The session context-menu enables the user to add new partitions to it from other stores on disk; similarly, the dataset context-menu enables adding new sessions to the dataset. The names of various partitions and sessions can be changed by the user via the \textsl{Properties} tab, and datasets can be saved and loaded from disk. Dataset files are persisted in an \textsl{Xml} format which facilitates manual editing.

A single session, deemed the \emph{active session}, can be visualized in \psistudio at any given time; the non-active sessions in the dataset are grayed out. The active session can be easily switched by double-clicking on a session, or by selecting the \textsl{Visualize} item in the session context-menu. 

Dataset objects can also be created and manipulated programmatically, and APIs that operate over datasets enable various data processing scenarios. For instance, developers can perform custom computations over the streaming data contained in a dataset, such as generating various statistics. One can also create derived partitions that contain new streams of data computed based on the streams in the existing partitions, e.g., processing a store of audio streams to create a derived partition containing extracted acoustic features. Such computations can also be encapsulated into \emph{batch processing task} functions, which can be dynamically loaded by \psistudio and surfaced for execution in the dataset and session context menus. Altogether, the dataset APIs, combined with the ability to visualize streams across different partitions and to switch between different sessions, facilitate data analytics, discovery and data-driven application tuning scenarios.

\subsection{Temporal Data Annotation}
\label{subsec:Tools_DataAnnotation}

Another important scenario in data-driven development is data annotation. \psistudio supports \emph{temporal data annotation} based on custom, developer-defined \emph{annotation schemas}. The temporal aspect refers to the fact that the tool supports segmentation, i.e., the ability to define start- and end-points for segments, and associate specific values with each segment based on predefined annotation schemas.

The annotation schemas can be defined programmatically and are persisted in a \texttt{.json} format, which allows for manual editing. The schema specification allows for using any desired \textsc{.NET} types as annotation values, and associating coloring and keyboard shortcuts with each value. The framework supports both \emph{unrestricted} annotation schemas, where the set of possible values is unlimited (e.g., useful when performing transcriptions), or \emph{finite} annotation schemas, where the set of possible values is finite and known ahead of time.

\psistudio provides various affordances for rapidly creating and editing such temporal annotations, leveraging mouse input and keyboard shortcuts. The annotations performed are persisted in \psif streams, and can therefore be processed just like any other automatically generated stream in various data analytics and tuning scenarios. This representation also enables semi-automatic annotation scenarios, where developers can automatically generate annotation streams, based for instance on existing heuristics or components, and then edit the constructed annotations manually to fine-tune them. For more in-depth information on how to perform annotations, we refer the interested readers to the online documentation \cite{PsiWebsite_Annotations}.

\section{Extensible Ecosystem of Components} 
\label{sec:Components}

In the previous two sections we have discussed the infrastructure and tools provided by the \psif framework. A third core aspect of the framework is its open, extensible ecosystem of component technologies, which promotes reuse and simplifies application development. In this section we review the existing array of components, as well as the framework's extensibility with respect to new data formats and visualizers.

\subsection{Stream Operators}

As we have already discussed in Section \ref{subsec:Runtime_StreamOperators}, \psif includes a built-in set of operators that work over generic streams of data. These stream operators are in fact implemented as simple components that can be customized in a variety of ways to perform a variety of tasks. The existing set includes: operators for transforming streams and selecting messages; operators for various mathematical and statistical computations; operators for data fusion, merging, sampling, and interpolation; operators for buffering and creating windows of messages; operators for generating streams of data; and so forth. In addition to simplifying and abstracting away the complexities involved in a number of common stream computations, these operators enable application developers to easily create connections between specialized, heterogeneous components with potentially mismatching interfaces. New operators can also be easily created. A comprehensive description of the current set of operators is beyond the scope of this paper and we refer the interested reader to the online documentation available in \cite{PsiWebsite_BasicStreamOperators}.

\subsection{Components}

In addition to the generic stream operators, \psif includes a growing set of components that implement and wrap a variety of multimodal sensing and processing technologies. 

A number of sensor components are available, allowing audio and visual data capture from microphones and cameras. In addition, components that provide access to specialized sensors, such as Kinect for Windows, Azure Kinect DK, and Intel RealSense are also available, as well as a number of ``file-sensor'' components that surface streams of data contained in various types of media files, such as CSV files or .mp4 videos. 

The framework provides a number of components for image and audio processing, including various image encoders and decoders, and an array of stream operators for basic image manipulation. In the audio space, the framework currently provides components for audio resampling and acoustic feature extraction. A number of speech, language, and computer vision components are also available: from components that wrap the \texttt{System.Speech} and \texttt{Microsoft.Speech} APIs and provide voice activity detection and speech recognition functionality locally, to components that run various cloud-based Azure Cognitive Services APIs for speech recognition, intent detection, chat, and computer vision. Finally, the framework also provides components for running ONNX deep-net models.

The components currently available in the GitHub repository are only meant to represent the seeds for what we hope will become a much broader ecosystem with community contributions. We have already described in sections \ref{subsec:Runtime_Components} and \ref{subsec:Runtime_CompositeComponents} the patterns for writing new components. As we have seen, the runtime architecture affords a straightforward model for component authoring, allowing developers to easily construct new components (simple or composite), while insulating them from the complexities of the concurrent execution environment in which these components will run.

\subsection{Extensibility}

The \psif framework was designed to be open and extensible in a number of ways. We have already mentioned the open ecosystem of components, as well as the capability to interop with other ecosystems and languages, such as Python or ROS (Section \ref{subsec:Runtime_RemotingInterop}). Additionally, \psif can be extended to include support for custom data formats and visualizers.

With respect to data, we have seen in Section \ref{subsubsec:Runtime_Persistence} that \psif provides a performant infrastructure for persisting data. The infrastructure relies on an efficient mechanism for binary serialization and is optimized for throughput. At the same time, legacy data may often come in many other formats. The data-read APIs in \psif were designed in a layered manner and a mechanism is provided to enable the creation of new importers for new data formats \cite{PsiWebsite_ThirdPartyStreamReaders}. The APIs therefore allow the construction and use of other streaming data sources, such as WAV and MPEG files, ROS bags, and so forth. Once a data importer is constructed, it can be easily used from the various higher-level APIs, e.g., for dataset processing, and by tools such as \psistudionospace, which can in this way visualize data stored in 3rd party formats.

Similarly, the visualization subsystem provides APIs and mechanisms that enable developers to write third party visualizers. Once written, these visualizers can be easily loaded and used as extensions in \psistudio \cite{PsiWebsite_ThirdPartyStreamReaders}.

\section{Related Work}
\label{sec:RelatedWork}

Platform for Situated Intelligence makes contributions in several areas, cutting across development process, tools, programming surface, and runtime behavior. The \psif runtime builds upon and extends well established models of parallel computation (Kahn process networks \cite{gilles1974semantics}) as well as methods for reasoning about time and causality (e.g., Lamport's logical time model \cite{lamport2019time}). The runtime provides efficient, low-latency, parallel execution on multi-core systems using ideas from early research in stream processing systems \cite{stephens1997survey} and dataflow programming \cite{johnston2004advances}, but also from systems targeting signal and video processing such as Microsoft DirectShow \cite{chatterjee1997microsoft}. 

Not surprisingly, \psif bears similarities to some of the modern streaming systems aimed at distributed event processing at cloud scale, such as TimeStream \cite{qian2013timestream}, MillWheel \cite{akidau2013millwheel}, Naiad \cite{murray2013naiad}, Trill \cite{chandramouli2014trill}, StreamScope \cite{lin2016streamscope} or Apache Flink \cite{katsifodimos2016apache}. In all these systems, the user constructs computation graphs (pipelines) by chaining built-in and user-defined stream-processing operators containing arbitrary, stateful computation kernels. While TimeStream, StreamScope, and MillWheel only support directed acyclic graphs, Naiad and Flink support loops and incremental computation. They all enable determinate execution through the use of watermarks and timestamps, which are usually included with message envelopes. However, their treatment of time varies greatly: while MillWheel, Naiad and Flink simply provide temporal join and windowing operators, leaving it to the user to decide when and how to use them, StreamScope goes further by enforcing the ordering of events delivered to multi-input nodes, while TimeStream converts all joins to temporal joins, by automatically restricting every join operation to events with overlapping time intervals. The \psif runtime on the other hand provides an explicit partial order of events based on originating-times, and enables time-based realignment of input streams through temporal data fusion operations. Moreover, since these other streaming systems were primarily designed to perform real-time analytics and data processing at scale, running on large clusters of commodity machines, they include design choices that favor fault tolerance, dynamic scaling, and reliable event delivery. The \psif runtime favors instead raw execution performance and ease of use. 

The ability to augment the computation graph at runtime, based on the data flowing through it, is a common feature in actor-based systems like Erlang \cite{armstrong2003making}, Akka \cite{akka}, and Orleans \cite{bernstein2014orleans}. Orleans provides support for streaming extensions, multiple types of channels with various levels of delivery guarantees, elastic scaling, and reliable execution (unlike Erlang and Akka which require user-provided supervisory trees). However, it has no built-in notion of time, does not provide a partial ordering of events, and does not guarantee FIFO behavior over some of its channels. Because it targets cloud applications, it makes similar design choices to the stream processing systems mentioned above. In \psifnospace, dynamic pipelines are enabled by the \texttt{Parallel()} operator, as discussed in Section \ref{subsec:Runtime_DynamicPipelines}.

The \psif programming model comes with a carefully designed API based on functional extensions to \textsc{C\#} and the \textsc{.NET} framework. It is closely related to LINQ \cite{meijer2006linq} and Reactive Extensions (RX) \cite{liberty2011programming}, which provide a convenient model for processing sequences but have limited parallel computation support. The other systems that target the \textsc{.NET} platform, TimeStream, Naiad, Trill and Orleans, have a similar API design approach, providing an object-oriented layer for implementing new operators and a functional-flavored layer for defining the computation graph. They include similar means for automatic serialization, leveraging the type reflection and code generation capabilities of the \textsc{.NET} platform. However, unlike \psifnospace, none of these systems provide allocation-free deserialization, nor do they support in-place cloning, which means their performance suffers when the streams contain non-primitive types. Unlike \psifnospace, they rely completely on \textsc{.NET}'s automatic memory management and make no special provisions for large objects like video frames.

The focus \psif has on quasi real-time applications processing heterogeneous sensor streams at the edge rather than in the cloud is shared by many robotics platforms \cite{kramer2007development}, most notably Microsoft Robotics Studio \cite{jackson2007microsoft} and ROS \cite{quigley2009ros}. Microsoft Robotics Studio is an advanced development environment based on the \textsc{.NET} framework. Besides providing an efficient concurrent runtime, it includes support for distributed execution, a library of sensor and actuation components, a simplified visual programming language and a full 3d simulator. However, its programming model, based on continuations, is difficult and overly prescriptive, and its distributed service model is complex and not particularly performant. 

ROS is similar in scope, and has a large and ever increasing ecosystem of components supporting virtually all commercial robotic platforms and many state-of-the-art algorithms and methods. In contrast to Microsoft Robotics Studio, ROS has focused primarily on the distributed computation aspect, embracing a decoupled actor model in which every component runs in its own process. While message flow is peer-to-peer with no particular coordination once a connection is established, ROS, like Orleans and Microsoft Robotics Studio, provides a topology-aware master node that helps establish the connections, while \psif leaves the management of the network topology to the application developer. The stated goals of ROS---``peer-to-peer, tools-based, multi-lingual and thin'' \cite{quigley2009ros}---encourage clear decoupling and separation of components, but have the side effect of favoring coarse componentization due to the high cost of communication.  In \psifnospace, nodes tend to be finer grained and reside in the same process, with efficient message scheduling handled by the \psif runtime. This in turn enables \psifnospace's delivery policy management, back-pressure, and throttling, as well as significantly better communication performance. ROS, Orleans and Microsoft Robotics Studio support async remote calls to other components, while \psif only supports streaming. While ROS provides support for timestamps in message headers, such information has not been pervasive until ROS2, with the underlying DDS protocol now assuming both \emph{source} and \emph{received} times, and providing configurable \emph{quality of service guarantees}. ROS2 also includes support for basic time synchronization. Finally, the \psif component ecosystem tends to focus on human-centric, multimodal, integrative-AI systems, while ROS and the other robotic platforms tend to focus on environment-centric sensing, control, motion planning, navigation, etc. 

Several other frameworks have been developed to address infrastructural needs and are in use across various research communities. For example, the Social Signal Interpretation (SSI) \cite{wagner2013social} framework enables real-time recognition of social signals by providing a stream processing runtime and a library of components for audio/video and HCI devices, stream processing, signal processing and machine learning. It distinguishes between high-frequency streams and low-frequency events, and provides different means to manipulate both. It provides means for synchronizing streams, but unlike \psifnospace, it assumes fixed stream rates. IrisTK \cite{skantze2012iristk} is a toolkit tailored for developing interactive dialog systems that operate in face-to-face, multiparty scenarios. The toolkit includes a set of modules for perception and production, and allows for authoring interaction flow via a variant of the Harel state-chart formalism \cite{harel1987statecharts}. 

MediaPipe \cite{lugaresi2019mediapipe} allows for creating custom multimodal perception pipelines, where sensor data such as audio and video is processed via a configurable set of connected computation nodes. The framework provides infrastructure for coordinating computation among nodes in the pipeline, including mechanisms for input synchronization and controlling flow on streams. It also provides basic tools for tracing message timings and visualizing the pipeline structure. In contrast to \psif (which is a managed, .NET framework), MediaPipe is built on native, C++ code, and is not supported on Windows.

As can be seen, \psif bears similarities to a number of related frameworks, infrastructures, and streaming systems that have been developed over the years, with important differences in implementation and scope. The primary goal in developing \psif has been to provide modern infrastructure, tools and components for rapidly authoring and tuning applications that process multimodal, temporally streaming data and integrate many different component technologies into end-to-end applications. However, as discussed in Section \ref{subsec:Runtime_RemotingInterop}, \psif is designed to be extensible, with capabilities to interop with other systems. Developers should ideally be able to easily create hybrid systems by bridging \psif to other frameworks as needed in order to create applications that combine the strengths of each infrastructure and ecosystem.

\section{Conclusion}
\label{sec:Conclusion}

We have introduced Platform for Situated Intelligence, or in short \psifnospace, an open-source\footnote{\url{https://github.com/microsoft/psi}} framework that supports and accelerates development and research for multimodal, integrative-AI systems. The framework leverages the affordances of the \textsc{.NET} managed programming environment, and includes a runtime, set of tools, and ecosystem of components to enable easy authoring and tuning of end-to-end applications.

At its base, the \psif \emph{runtime} provides a performant infrastructure for working with multimodal, temporally streaming data. \psif implements a publish-subscribe architecture, where computation is performed in a graph of components connected via streams of data. The \psif streaming infrastructure provides support for deeply reasoning about time and latency, and for operations like data synchronization, interpolation, and fusion. \psif also provides for high efficiency, high throughput logging, and data replay, enabling a variety of data-driven application development and tuning scenarios. The execution model implemented by the \psif runtime aims to maximize resource utilization by leveraging concurrent execution and pipeline parallelism, while enabling graceful performance degradation under load. At the same time, it insulates component developers from the challenges that often arise in concurrent execution environments.

The framework provides a set of \emph{tools} that support application development, debugging, maintenance and tuning. Among these, Platform for Situated Intelligence Studio, or in short \psistudionospace, provides for multimodal data visualization, annotation and processing. The tool allows developers to construct complex visualization layouts to inspect data persisted by \psif applications (or contained in other streaming formats), both offline and live. \psistudio also enables visualizing the structure of the application pipeline and related stream- or component-level diagnostic information, which can speed up debugging and tuning. The tool enables temporal annotation scenarios using user-defined annotation schemas, and enables developers to work with datasets and perform custom batch processing and data analytics tasks. The data processing capabilities are also made available via a companion command-line tool.

Finally, the third aspect of \psif is its extensible array of \emph{components} that encapsulate various technologies and foster fast prototyping via reuse. The components currently available in the GitHub repository center on multimodal sensing and processing: they include a variety of sensor components, e.g. for cameras, microphones, as well as other sensors such as Azure Kinect, audio and image processing components, speech and language components, as well as components that wrap various cloud services, and components that allow for running existing ML models. The current components are the seeds of what we hope will become, via community contributions, a growing ecosystem of components.

As sensing and perception technologies continue to improve, exciting opportunities are arising to build the next generation of AI systems that can reason about and/or act in the open world. We believe \psif can significantly accelerate research and development in this space, where applications must harness the power of multiple AI technologies to process multimodal temporal data streams in real-time. We encourage the community to use the framework, and bring contributions on all fronts: from filing bugs and raising issues, all the way to developing new features, and extending the set of components and tools. We hope that, with community engagement, the framework will lower the barrier to entry in this space and foster more research in multimodal, integrative-AI systems.

\section*{Acknowledgments}
Many people have contributed to the development of the Platform for Situated Intelligence framework, from designing and implementing the runtime, tools, and components, to testing and providing feedback. We would like to acknowledge the contributions by Mihai Jalobeanu on core aspects of the runtime, including time-synchronization, allocation-free deep cloning and serialization, memory pools, persistence and remoting via memory-mapped files, the scheduler, and the fluent operator-based API. We would like to also thank all of our present and past team members, collaborators, and early adopters at Microsoft Research, including Debadeepta Dey, John Elliott, Don Gillett, Daniel McDuff, Kael Rowan, Lev Nachmanson and Mike Barnett. Finally, we would like to thank our external collaborators and the GitHub community for their very valuable feedback, discussions, and contributions.

\clearpage

\bibliographystyle{plain}

\end{document}